\documentclass[10pt,twocolumn,letterpaper]{article}

\usepackage[pagenumbers]{cvpr} % To force page numbers, e.g. for an arXiv version

\usepackage{graphicx}
\usepackage{amsmath}
\usepackage{amssymb}
\usepackage{booktabs}
\usepackage{soul}
\usepackage{multirow}

\usepackage[pagebackref,breaklinks,colorlinks]{hyperref}

\usepackage[capitalize]{cleveref}
\crefname{section}{Sec.}{Secs.}
\Crefname{section}{Section}{Sections}
\Crefname{table}{Table}{Tables}
\crefname{table}{Tab.}{Tabs.}

\def\cvprPaperID{1307} % *** Enter the CVPR Paper ID here
\def\confName{CVPR}
\def\confYear{2022}

\begin{document}

\title{Tree Energy Loss: Towards Sparsely Annotated Semantic Segmentation}

% \author{Zhiyuan Liang\\
% Beijing Institute of technology\\
% {\tt\small liangzhiyuan@bit.edu.cn}
% \and
% Tiancai Wang \qquad Xiangyu Zhang \qquad Jian Sun\\
% Megvii Technology\\
% % First line of institution2 address\\
% {\tt\small \{wangtiancai,zhangxiangyu,sunjian\}@megvii.com}
% \and
% Jianbing Shen\\
% Beijing Institute of technology\\
% % First line of institution address\\
% {\tt\small shenjianbing@bit.edu.cn}
% }
\newcommand{\sep}{\quad}

\author{Zhiyuan Liang$^{1}$\thanks{This work was performed during an internship at MEGVII Technology. This work was supported by The National Key Research and Development Program of China (2020AAA0105200) and Beijing Academy of Artificial Intelligence (BAAI). Corresponding author: \textit{Jianbing Shen}. Email:  shenjianbingcg@gmail.com}  \qquad Tiancai Wang$^{2}$ \qquad Xiangyu Zhang$^{2}$ \\
\qquad Jian Sun$^{2}$  \qquad Jianbing Shen$^{1}$\vspace{1mm}\\
$^1$Beijing Institute of Technology \sep
$^2$MEGVII Technology
}

\maketitle

%%%%%%%%% ABSTRACT
\begin{abstract}  
Sparsely annotated semantic segmentation (SASS) aims to train a segmentation network with coarse-grained (i.e., point-, scribble-, and block-wise) supervisions, where only a small proportion of pixels are labeled in each image.
In this paper, we propose a novel tree energy loss for SASS by providing semantic guidance for unlabeled pixels. 
% The tree energy loss builds the minimum spanning trees on original images and intermediate features to model the low-level and high-level pair-wise affinities, respectively.
The tree energy loss represents images as minimum spanning trees to model both low-level and high-level pair-wise affinities.
By sequentially applying these affinities to the network prediction, soft pseudo labels for unlabeled pixels are generated in a coarse-to-fine manner, achieving dynamic online self-training. 
% The tree energy loss represents images as minimum spanning trees and models the pair-wise affinity via the low-level visual and high-level semantic information. 
% By merging the network predictions with multi-level affinities sequentially, soft pseudo labels for unlabeled pixels are generated in a coarse-to-fine manner, resulting in the dynamical online self-training during the training phase. 
% Tree energy loss leverages the inherent multi-level structural information of images to conduct the online self-training between the labeled and unlabeled regions. 
% The tree energy loss is easy and effective to be integrated into many traditional semantic segmentation frameworks.
The tree energy loss is effective and easy to be incorporated into existing frameworks by combining it with a traditional segmentation loss.
% By combining the tree energy loss with the traditional segmentation loss, 
% a clear and effective training  framework is proposed for SASS tasks.
Compared with previous SASS methods, our method requires no multi-stage training strategies, alternating optimization procedures, additional supervised data, or time-consuming post-processing while outperforming them in all SASS settings.
% It can implicitly generate pseudo-labels for unlabeled areas and deal with SASS in a one-stage manner.
% It is easy and effective to be integrated into many existing segmentation frameworks. Experimental results demonstrate that tree energy loss outperforms other methods under block-wise, scribble-wise, and point-wise annotated settings. 
Code is available at \url{https://github.com/megvii-research/TreeEnergyLoss}.
\end{abstract}

%%%%%%%%% BODY TEXT
\section{Introduction} \label{sec:intro}

\begin{figure}[t]
	\begin{center}
		\includegraphics[width=0.9\linewidth]{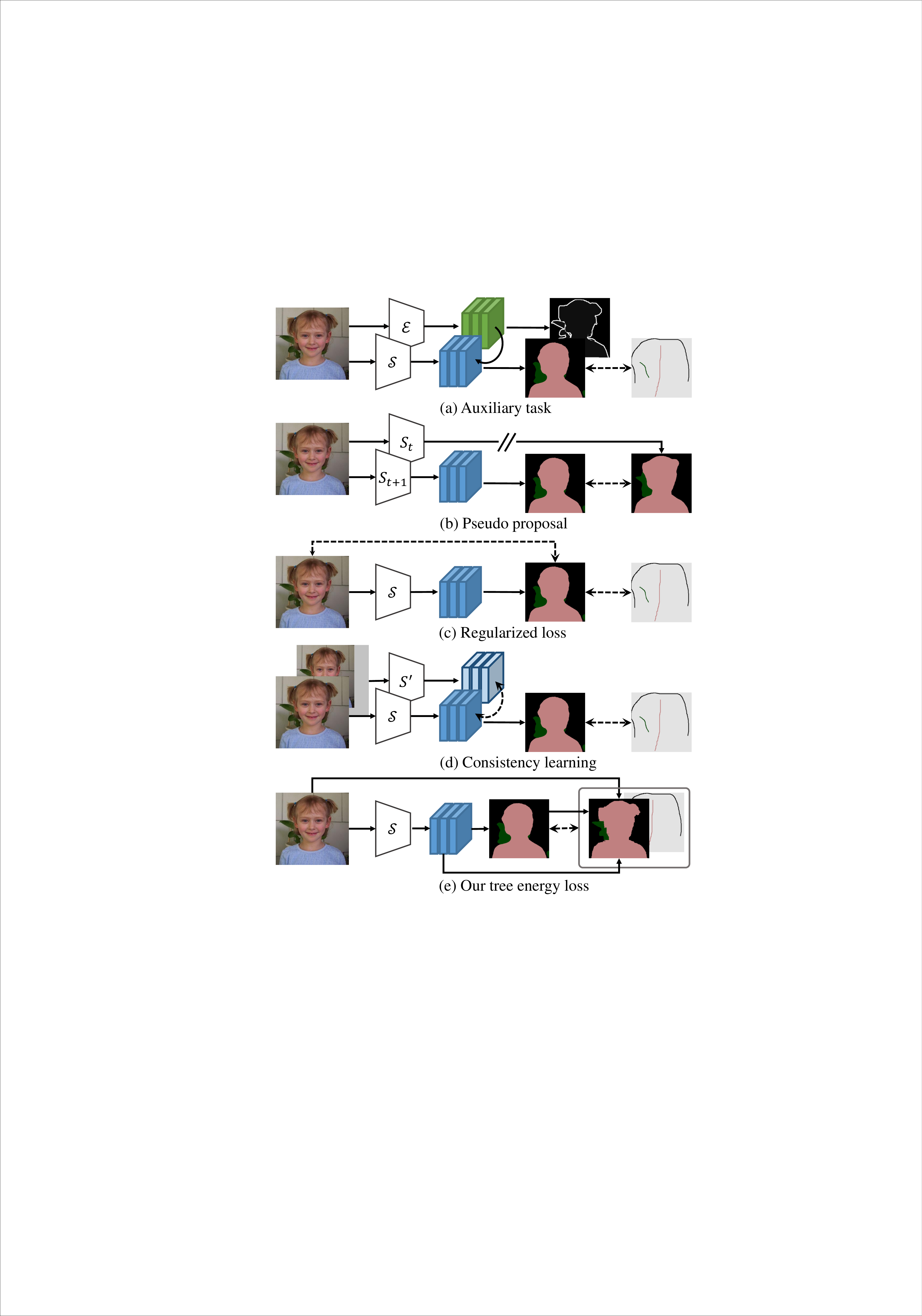}
	\end{center}
	\vspace{-0.65cm}
	\caption{Illustration of current SASS approaches. $\mathcal{S}$ and $\mathcal{E}$ denote the segmentation and auxiliary models, respectively. Our method leverages the minimum spanning trees (MSTs) to capture both low-level and high-level affinities to generate soft pseudo labels, performing online self-training.}
% 	Our method can capture the structure information of the image, which achieves higher edge accuracy in the unlabeled area. Significant differences are highlighted within the bounding boxes.}
	\vspace{-4mm}
	\label{Fig:SASS_methods}
\end{figure}

\begin{figure*}[t]
	\begin{center}
		\includegraphics[width=1.\linewidth]{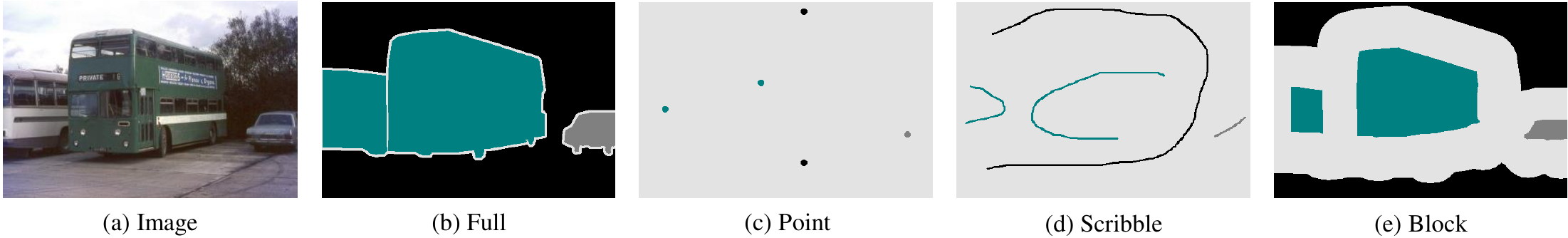}
	\end{center}
	\vspace{-0.5cm}
	\caption{Different types of training annotations for semantic segmentation. The background class is annotated in black.}
	\vspace{-2mm}
	\label{Fig:annotations}
\end{figure*}

%% task definition
Semantic segmentation, aiming to assign each pixel a semantic label for given images, is one of the fundamental tasks in computer vision. Previous methods \cite{chen2017deeplab, liu2019auto, song2019learnable, song2020rethinking, wang2021exploring} tend to leverage large amounts of fully annotated labels like Fig.~\ref{Fig:annotations}(b) to achieve satisfying performance. However, manually annotating such high-quality labels is labor-intensive. 
To reduce the annotation cost and preserve the segmentation performance, some recent works research on semantic segmentation with sparse annotations, such as point-wise \cite{bearman2016s} and scribble-wise ones \cite{lin2016scribblesup}. As shown in Fig.~\ref{Fig:annotations}(c-d), the point-wise annotation assigns each semantic object with a single-pixel label while the scribble-wise annotation draws at least a scribble label for the object. 

As illustrated in Fig.~\ref{Fig:SASS_methods}(a-d), existing approaches are mainly based on \textit{auxiliary tasks}, \textit{pseudo proposals}, \textit{regularized losses}, and \textit{consistency learning} to solve SASS. 
However, there are some shortcomings in these approaches. 
The predictive error from the auxiliary task \cite{vernaza2017learning, wang2019boundary, lee2021weakly} may hinder the performance of semantic segmentation. 
The proposal generation \cite{lin2016scribblesup, zhang2021affinity, xu2021scribble} is time-consuming and usually calls for a multi-stage training strategy.
The regularized losses \cite{tang2018normalized,tang2018regularized,marin2019beyond,obukhov2019gated,tian2021boxinst} ignore the domain gap between the visual information and the high-level semantics,
% while the consistency learning \cite{pan2021scribble, qian2019weakly, ke2021universal, chen2021seminar} conducted on intermediate features, fails to supervise the unlabeled pixels at category level.
and the consistency learning \cite{pan2021scribble, qian2019weakly, ke2021universal, chen2021seminar, zhou2021group, zhou2022rsca} fails to directly supervise the unlabeled pixels at the category level.
In this paper, we aim to alleviate these shortcomings and introduce a simple yet effective solution.

In SASS, each image can be divided into labeled and unlabeled regions. The labeled region can be directly supervised by the ground truth, 
% Therefore, how to leverage the labeled data for the supervised learning of unlabeled regions is an open question. 
% \st{Therefore, how to assist the learning of the unlabeled region with the labeled data is an open question.}
while how to learn from the unlabeled region is an open question.
For the region of the same object, the labeled and unlabeled pixels share similar patterns on low-level color (RGB value of image) and high-level responses (features produced by CNN).
% For one object in image, the labeled and unlabeled regions share similar patterns on low-level color (RGB value of the pixel) and high-level responses (features produced by CNN). 
% \st{Thus, how to make full use of such similarity prior is the key for SASS.}
Utilizing such similarity prior in SASS is intuitive. 
% The way to establish contact between labeled and unlabeled pixels is significant to SASS.
Inspired by the tree filter \cite{bao2013tree,yang2014stereo}, which can model the pair-wise similarity with its structure-preserving property, 
% we leverage this property to achieve the consistent supervision for the unlabeled regions.
we leverage this property to generate soft pseudo labels for unlabeled regions and achieve online self-training.

Specifically, we introduce a novel tree energy loss (TEL) based on the low-level and the high-level similarities of image (see Fig.~\ref{Fig:SASS_methods}(e)). 
% In SASS, each image consists of both labeled and unlabeled regions, which constrains the effectiveness of the per-pixel classification loss for model learning. 
% Thus we resort to the pair-wise relation between pixels to boost the performance.
% We propose to boost the performance with the pair-wise relation between pixels.
% However, it is non-trivial to model the pair-wise relation since existing popular approaches like non-local operation \cite{wang2018non} may capture inaccurate long-range relation due to the sparse supervision.
% On the other hand, modeling the relation between adjacent pixels is much easier.
% Inspired by tree filter \cite{bao2013tree,yang2014stereo,song2019learnable}, which first models the short-range relations between adjacent pixels then gradually aggregates them to form the long-range ones, we propose a novel tree energy loss for SASS. 
% In SASS, each image consists of two parts: labeled and unlabeled regions. 
% How to leverage the labeled and unlabeled data for model training is an open question.
% Inspired by tree filter \cite{bao2013tree,yang2014stereo,song2019learnable}, we propose a novel tree energy loss for SASS. 
In TEL, two minimum spanning trees (MSTs) are built on the low-level color and the high-level semantic features, respectively.
Each MST is obtained by sequentially eliminating connections between adjacent pixels with large dissimilarity, so less related pixels are separated and the essential relation among pixels is preserved. 
% Since the sparse annotation tends to mark out different objects in the image, performing online self-training based on the MST is reasonable. 
Then, two structure-aware affinity matrices obtained by accumulating the edge weights along the MSTs are multiplied with the network predictions in a cascading manner, producing soft pseudo labels.
Finally, the generated pseudo labels are assigned to the unlabeled regions. Combining the TEL with a standard segmentation loss (e.g., cross-entropy loss), any segmentation network can learn extra knowledge from unlabeled regions via dynamic online self-training.
% The proposed tree energy loss can be regarded as a hybrid approach of pseudo proposals and the regular loss, but further be extended to high-level consistency.
% and any segmentation network can learn extra knowledge from the whole image.

% The other shortcoming for most of previous SASS approaches is the complex implementation including multi-stage training strategies, alternating optimization procedures, additional supervised data, and the time-consuming post-processing. It prevents them from being widely used. By contrast, the tree energy loss is simple and effective. It can serve as a plug-and-play module for most of existing segmentation network, making the network trained in a single-stage manner. During the inference time, it can be removed to avoid extra computational costs.

To comprehensively validate the effectiveness of TEL, we further enrich the SASS scenarios by introducing a block-wise annotation setting (see Fig.~\ref{Fig:annotations}(e)), where the amount of annotations is located between the full and scribble settings.
% The block-wise annotations, whose annotation amounts are located between the full and scribble annotations. 
In this way, we can grade the SASS into three levels, i.e., point, scribble, and block. 
Experimental results show that TEL can significantly boost segmentation performance without introducing extra computational costs during inference. 
Equipped with recent segmentation networks, our method can achieve state-of-the-art performance under various annotated settings.

The main contributions are summarized as follow. We propose a novel tree energy loss (TEL) for SASS. TEL leverages minimum spanning trees to model the low-level and high-level structural relation among pixels. A cascaded filtering operation is further introduced to dynamically generate soft pseudo labels from network predictions in a coarse-to-fine way. TEL is clean and easy to be plugged into most existing segmentation networks. For comprehensive validation, we further introduce a block-annotated setting for SASS. Our method outperforms the state-of-the-arts under the point-, scribble- and block-annotated settings. 

%-------------------------------------------------------------------------
\begin{figure*}[t] 
	\begin{center}
		\includegraphics[width=0.9\linewidth]{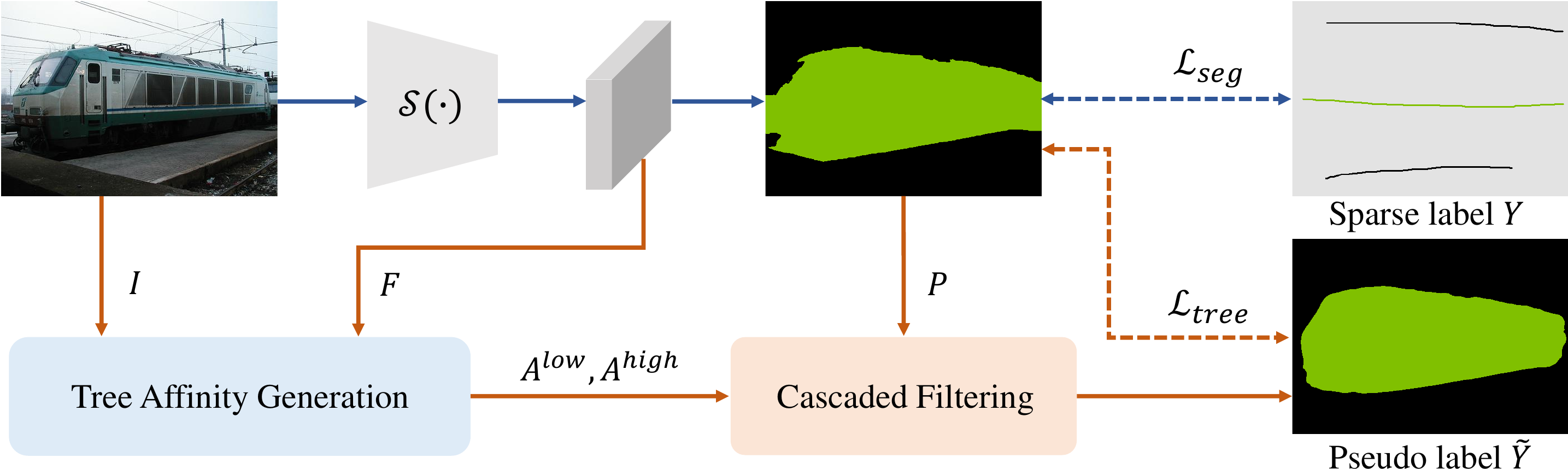}
	\end{center}
	\vspace{-6mm}
	\caption{Flowchart of the proposed single-stage SASS method, which is realized by incorporating an \textcolor[RGB]{197,90,17}{auxiliary branch} into the \textcolor[RGB]{47,85,151}{traditional segmentation model} $\mathcal{S}(\cdot)$.
	During training, the predicted masks $P$ are split into the labeled and unlabeled parts, which are supervised by the segmentation loss $\mathcal{L}_{seg}$ and the tree energy loss $\mathcal{L}_{tree}$, respectively.
	To obtain pseudo labels for unlabeled pixels, the Tree Affinity Generation procedure (Eqs.~\ref{eq:edge_weight}-\ref{eq:affinity_projection}) 
	first utilizes the color information $I$ and semantic features $F$ to generate the low-level and high-level affinity matrices $A^{low}, A^{high}$. 
	Then the Cascaded Filtering operation (Eqs.~\ref{eq:treeloss_aggregation}-\ref{eq:treeloss_filtering})
	converts the network predictions $P$ into soft pseudo labels $\tilde{Y}$. During testing, the \textcolor[RGB]{197,90,17}{auxiliary branch} is removed to avoid extra computational costs.}
	\vspace{-2mm}
	\label{Fig:flowchart}
\end{figure*}

\section{Related Works}  
\noindent \textbf{Sparsely Annotated Semantic Segmentation:} 
% \ul{(double check the tense)} 
Sparsely annotated semantic segmentation aims to train the segmentation model with coarse-grained annotated data. Previous works mainly focus on the point-level and the scribble-level supervisions. 
% What's the Point \cite{bearman2016s} first presents the semantic segmentation task with point annotations. It combines image-level and point-level supervisions into the loss function and leverages the objectness prior to improve the performance. 
What's the Point \cite{bearman2016s} first presents the semantic segmentation task with point annotations. It combines the objectness prior, the image-level supervision, and point-level supervision into the loss function.
PDML \cite{qian2019weakly} proposes the point-based distance metric learning to model the intra- and inter-category relations across images. WeClick \cite{liu2021weclick} utilizes temporal information of the video sequence and distills the semantic knowledge from a more complex model.
Seminar \cite{chen2021seminar} introduces seminar learning through EMA-based teacher models. 
% However, the performance of these methods is far from state-of-the-art. 
To narrow the performance gap with fully annotated methods,
an increasing number of scribble-annotated semantic segmentation methods have appeared.
ScribbleSup \cite{lin2016scribblesup} constructs a graphical model to alternatively propagate the scribble annotations and learn the model parameters. 
RAWKS \cite{vernaza2017learning} and BPG \cite{wang2019boundary} adopt edge detectors to progressively refine the predictions for sharper semantic boundaries.
% SPML \cite{ke2021universal} proposes a single pixel-wise metric learning framework for different types of weakly supervised semantic segmentation.
A\textsuperscript{2}GNN \cite{zhang2021affinity} blends the multi-level supervision and solves the segmentation problem with graph neural networks.
PSI \cite{xu2021scribble} utilizes multi-stage semantic features to progressively infer the predictions and pseudo labels. 
URSS \cite{pan2021scribble} learns to reduce the uncertainty of the segmentation model by random walks, coupled with a self-supervised learning strategy.
To capture the relation between labeled and unlabeled pixels, a variety of regularized losses \cite{tang2018normalized,tang2018regularized,marin2019beyond,obukhov2019gated} are proposed. These methods use the low-level (i,e., spatial and color) information  of images and train the model in two stages. In the first stage, the segmentation model is just trained with segmentation loss. Then a regularized loss is further adopted to fine-tune the model in the second stage.
% can be applied in existing segmentation frameworks. 
% However, their performance is far from state-of-the-art.
% This paper presents a tree energy loss for SASS. The tree energy loss generates soft labels for unlabeled pixels based on both the low-level and the high-level structural information and conducts online self-training.

\noindent \textbf{Tree Filter:}
Modeling pair-wise relation is significant for many computer vision tasks. Regarding an image as an undirected planar graph, where nodes are all pixels and edges between adjacent nodes are weighted by the appearance dissimilarity, the minimum spanning tree (MST) can be constructed by removing edges according to substantial weights. Since the gradient between adjacent pixels can be viewed as the intensity of object boundaries, the nodes tend to interact with each other preferentially within the same object on the tree. 
Due to the structure-preserving property of MST, the traditional tree filters are applied in stereo matching \cite{yang2012non,yang2014stereo}, salient object detection \cite{tu2016real}, image smoothing \cite{bao2013tree}, denoising \cite{stawiaski2009minimum}, and abstraction \cite{koga2011structural}.
Recently, LTF \cite{song2019learnable} presents a learnable tree filter to capture the long-range dependencies for semantic segmentation. LTF-V2 \cite{song2020rethinking} combines the learnable tree filter and the Markov Random Field \cite{li1994markov} to further improve the performance. 
% Our method is inspired by these works which appy the tree filter to original images and intermediate features,
% Our method is inspired by these works above but for a different purpose. Instead of applying tree filter to original images or intermediate features, we capture both low-level and high-level affinities and apply tree filter to network predictions for pseudo label generation, leading to dynamic online self-training.
% Since the category label is partially missed in SASS, we construct MSTs based on low-level color and high-level semantic features while carrying out the filter operation at the category level, leading to dynamic online self-training during training.
% Since the label is partially missed in the SASS task, we build a graph model based on low-level image information and highand conduct online self-training along the MST.  

\section{Methodology}
% In this section, 
% We first formulate the motivation of SASS~\S\ref{Sec:problem_formulation}.
In this section, we first emphasize our motivation in Sec.~\ref{Sec:problem_formulation}.
Then the overall architecture combining the traditional segmentation loss with the proposed tree energy loss (TEL) is introduced in Sec.~\ref{Sec:total_loss}.
After that, we describe the details of TEL in Sec.~\ref{Sec:tree_loss}. Finally, we discuss the main differences from previous related works in Sec.~\ref{sec:diss}.

% In this section, we elaborate on the proposed tree energy loss for Sparsely Annotated Semantic Segmentation (SASS). 
% We first formulate the problem of SASS~\S\ref{Sec:problem_formulation}, 
% then present an unified solution of SASS~\S\ref{Sec:total_loss}.
% Finally, the tree energy loss is proposed to solve the SASS problem~\S\ref{Sec:tree_loss}. 
% Then the tree energy loss is proposed to solve the SASS problem~\S\ref{Sec:tree_loss}. 
% Finally, an efficient single-stage framework (shown in Fig.~\ref{Fig:flowchart}) is introduced by combining the traditional segmentation loss with our tree energy loss
% Finally, an efficient single-stage framework with our tree energy loss is introduced by combining the traditional segmentation loss ~\S\ref{Sec:total_loss}.
% In this section, we elaborate on the proposed tree energy loss for Sparsely Annotated Semantic Segmentation (SASS). The flowchart is illustrated in Fig.~\ref{Fig:flowchart}. We first formulate the problem of SASS and propose to solve SASS in a semi-supervised manner~\S\ref{Sec:problem_formulation}. Then the tree energy loss is introduced to propagate soft labels for unlabeled data~\S\ref{Sec:tree_loss}. Finally, an efficient single-stage SASS training framework is proposed by combining the traditional segmentation loss with our tree energy loss~\S\ref{Sec:total_loss}.

\subsection{Motivation}\label{Sec:problem_formulation}
The SASS task aims to train a dense prediction model with coarse-grained (i.e., point-, scribble- or block-wise) labels, where the annotations of most pixels are invisible during training.
In SASS, the whole image can be separated into two parts: labeled set $\Omega_L$ and unlabeled set $\Omega_U$. 
For the labeled set $\Omega_L$, one can simply use the corresponding ground truth for training. As for the $\Omega_U$, it tends to be ignored in the traditional semantic segmentation framework, resulting in performance degradation. 
% Recently, some methods have been proposed to alleviate this issue. 
% However, these methods are either ignore the multi-level visual information of image or too complex to be easily used. 
% However, these methods usually are of complex design, requiring multi-stage optimization and their performance are far from satisfactory.
% \st{To take full use of unlabeled data, we propose an effective online self-training strategy for SASS by considering the pixel-wise consistency and the informative complementarity of the low-level and high-level features.}
% To address this issue, we try to provide extra supervision for unlabeled data by modeling the pair-wise relation of images.
% This paper aims to take full use of unlabeled data to boost the performance of the segmentation model.
This paper aims to present a simple yet effective solution for SASS.
Since pixels belonging to the same object share similar patterns at different feature levels, we leverage these similarities to provide the additional supervision for unlabeled pixels in $\Omega_U$.
Inspired by tree filter \cite{bao2013tree,yang2014stereo,song2019learnable}, we model such pair-wise similarity based on its structure-preserving property. The pair-wise similarity together with the network prediction is used to generate soft pseudo labels for unlabeled pixels. Cooperated with the supervised learning in $\Omega_L$, an online self-training framework is constructed, achieving the progressive improvement of both network predictions and pseudo labels during training.

\subsection{Overall Architecture}\label{Sec:total_loss}
% The unified single-stage training flowchart is illustrated in Fig.~\ref{Fig:flowchart}. 
% \textcolor{red}{Maybe we should modify the section title with "Overall architecture" and give some description about Fig.3.} 
% Fig.~\ref{Fig:flowchart} illustrates the overall architecture of our online self-training flowchart, which consists of a semantic segmentation branch for labeled data and an auxiliary branch for unlabeled pixels. To obtain the pseudo labels for unlabeled pixels, the low-level and high-level MSTs are first built upon the original images $I$ and the embedded features $F$ to  calculate the corresponding pair-wise affinity matrices. Then these affinity matrices are served as filter kernels and serially applied to the network predictions to generate high-quality pseudo labels. 
Fig.~\ref{Fig:flowchart} illustrates the overall architecture of our method, which is composed of a segmentation branch for labeled pixels and an auxiliary branch for unlabeled pixels. The segmentation branch assigns the sparsely annotated label $Y$ to the labeled pixels. 
For the auxiliary branch, the pair-wise affinity matrices $A^{low}$, $A^{high}$ are generated from the original image $I$ and the embedded features $F$. Then the affinity matrices $A^{low}$, $A^{high}$ are used to refine the network prediction $P$ and generate soft pseudo label $\tilde{Y}$. 
The soft labels generated are assigned to the unlabeled pixels. 
% First, the low-level and high-level MSTs are built upon the original images and the embedded features to calculate the corresponding pair-wise affinity matrices.
% Then, these affinity matrices are served as filer kernels and
% serially applied to the network predictions to generate high-quality pseudo labels. 
% Finally, the soft label assignment is conducted between the pseudo labels and the original predictions, which can provide additional category supervision for unlabeled regions.
Therefore, the overall loss function includes a segmentation loss $\mathcal{L}_{seg}$  and a tree energy loss $\mathcal{L}_{tree}$,
\begin{equation}
\mathcal{L} = \mathcal{L}_{seg} + \lambda \mathcal{L}_{tree},
\label{eq:total_loss}
\end{equation}
where $\lambda$ is a balance factor for two losses. By leveraging two losses jointly, complementary knowledge can be learned by the whole segmentation network.
% The proposed tree energy loss learns to propagate soft labels for unlabeled data dynamically. 
% For the labeled pixels, a standard segmentation loss is adopted.
% we use the standard segmentation loss for training.
For the $\mathcal{L}_{seg}$, we follow previous works \cite{tang2018normalized,tang2018regularized,marin2019beyond,obukhov2019gated,liu2021weclick} and formulate it as a partial cross-entropy loss: 
\begin{equation}
\mathcal{L}_{seg} =  -\frac{1}{|\Omega_L|}\sum_{\forall{i}\in\Omega_L}Y_i \log(P_i),
\label{eq:seg_loss}
\end{equation}
where $P_i$ and $Y_i$ are the network prediction and the corresponding ground truth at the location $i$. 
% $\Omega_L$ is the labeled set of the image.
% \textcolor{red}{Since the segmentation loss only supervises the model training with labeled samples, the feature distribution of unlabeled samples can be of arbitrary form.} 
% Thus the tree energy loss is introduced to refine the feature distribution with pair-wise affinity. 
% For the unlabeled pixels, a novel tree energy loss is designed, which will be presented in the next section.
As for the proposed $\mathcal{L}_{tree}$, it will be presented in the next section.

\begin{figure}[t]
	\begin{center}
		\includegraphics[width=1.0\linewidth]{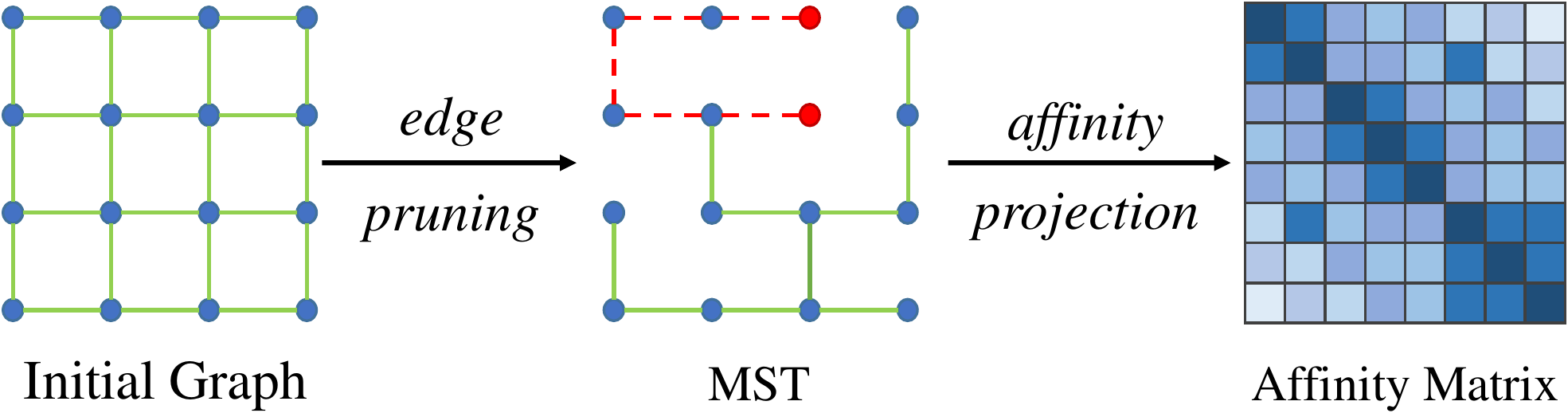}
	\end{center}
	\vspace{-6mm}
	\caption{The process of Tree Affinity Generation. An initial graph is first built on the given low-level color or high-level semantic features, then the MST is obtained by the \textit{edge pruning} algorithm \cite{gallager1983distributed}. On the MST, the distance between two vertices is calculated by the sum of edge weights along their hyper-edge. An example is illustrated in red dashed lines. Finally, the \textit{affinity projection} is conducted to project the distance map into an affinity matrix.}
	\vspace{-4mm}
	\label{Fig:affinity_gen} 
\end{figure}

\subsection{Tree Energy Loss}\label{Sec:tree_loss}
% Inspired by the tree filters in image processing, we propose the tree energy loss for the SASS problem. 
% The tree energy loss is proposed to leverage the unlabeled data by capturing multi-level image information during training. 
Given the training images with sparse annotations, TEL learns to provide category guidance for unlabeled pixels.
% generate soft pseudo labels dynamically. 
% \textcolor{red}{learns to propagate the category information dynamically}. 
The TEL mainly includes the following three steps: 
(1)~A tree affinity generation step to model the pair-wise relation.
(2)~A cascaded filtering step to generate pseudo labels.
(3)~A soft label assignment step to assign pseudo labels for unlabeled pixels.
Here, we will introduce TEL in detail. 
% First, the low-level and the high-level Minimum Spanning Trees (MSTs) are constructed and the corresponding pair-wise affinity matrices are calculated. 

% \textcolor{red}{Too long, we can simplify this part or put some contents to Sec.3.2.}
% The tree energy loss can capture multi-level structural information of images and provide additional category supervision for unlabeled regions.

% First, two types of minimum spanning tree (MST) are generated based on the low-level priors the high-level semantics of the image, respectively. Then, the pair-wise affinity matrices are produced according to the MSTs and serially merged with the prediction of the segmentation network, generating the pseudo labels for unlabeled pixels. 
% Finally, the soft label assignment is conducted between the pseudo labels and the original predictions. 
% Finally, the soft label assignment is conducted between the pseudo labels and the original predictions, leading to the interaction among pixels in whole training set $\Omega$.

% We will next describe the loss in detail. 
% Since the adjacent pixels with similar color information tends to have the same category, 
% The proposed tree loss is inspired by the tree filter. 
% The tree energy loss is proposed to aggregate and distribute the semantic information based on the structural relation among pixels.

% \noindent \textbf{Minimum Spanning Tree Construction.}
\noindent \textbf{Tree Affinity Generation.}
An image can be represented as an undirected graph $G=(V,E)$, where the vertice set $V$ consists of all pixels and the edges between two adjacent vertices make up the edge set $E$. 
As shown in Fig.~\ref{Fig:affinity_gen}, we adopt the architecture of a 4-connected planar graph, where each pixel is adjacent to up to 4 neighboring ones.
Let the vertice $i$ and vertice $j$ be adjacent on the graph, the low-level and high-level weight functions between them can be respectively defined as
\begin{equation}
\begin{aligned}
& \omega^{low}_{i,j} = \omega^{low}_{j,i} = {|I(i)-I(j)|}^2, \\
& \omega^{high}_{i,j} = \omega^{high}_{j,i} = {|F(i)-F(j)|}^2, 
\end{aligned}
\label{eq:edge_weight}
\end{equation}
% where $I(i) \in \mathbb{R}^{3\times h \times w}$ is the RGB color embedding
% and $F(i) \in \mathbb{R}^{256\times h \times w}$ is the semantic feature embedding of pixel $i$, respectively. 
where $I(i) \in \mathbb{R}^{3\times h \times w}$
and $F(i) \in \mathbb{R}^{256\times h \times w}$ are the RGB color and the semantic features of pixel $i$, respectively. 
$h$ and $w$ are the height and width of the downsampled input image. 
% The superscripts $l$ and $h$ denote the low-level and high-level information, respectively.
% \textcolor{red}{In detail, $I(i)$ is the RGB color feature of the vertice $i$ while 
% By feeding the features produced before the last classification layer of the segmentation model into a cascaded learnable $1 \times 1$ convolution,
% BatchNorm \cite{ioffe2015batch} and ReLU activation \cite{nair2010rectified} module, the $F(i)$ can be obtained. 
% \textcolor{red}{$F(i)$ can be generated by feeding the features, which are produced before the last classification layer of the segmentation model,
% into a cascaded $1 \times 1$ convolution, BatchNorm \cite{ioffe2015batch} and ReLU activation \cite{nair2010rectified} module.}
$F(i)$ is produced by a $1 \times 1$ convolutional layer, from the features before the classification layer of the segmentation model.
% The $F(i)$ can be produced by a learnable $1 \times 1$ convolutional layer, followed by a BatchNorm \cite{ioffe2015batch} and a ReLU activation \cite{nair2010rectified} layers.
% $F(i)$ can be produced by the Feature Embedding Module (FEM), which is implemented by a single $1 \times 1$ convolutional layer.
Once obtained the edge weights, a MST can be constructed by sequentially removing the edge with the largest weight from $E$ while ensuring the connectivity of the graph.
% The edges with relatively small weights are preserved and the similar pixels will interact with each other preferentially.
% \st{As shown in Fig.}~\ref{Fig:affinity_gen}, \st{we begin with the architecture of the traditional 4-connected planar graph, where each pixel is adjacent to up to 4 neighboring ones.
% After constructing the initial graph,} \textcolor{red}{What is the relation between the initial graph and the weight function in (3)?}
% in low-level color space and high-level feature space separately, 
% \st{the MST can be obtained by sequentially removing edges with large weights from $E$. The edges with relatively small weights are preserved and the similar pixels will interact with each other preferentially.} 
We construct both the low-level and the high-level MSTs with the \textit{Bor$\mathring{u}$vka} algorithm \cite{gallager1983distributed}.
% Notice that 
Based on the topology of MST, vertices within the same object share similar feature representations and tend to interact with each other preferentially.
% By constructing the MST, spatially closed pixels with less dissimilarity will be separated from each other. 
% Although the domain gap exists between low-level/high-level features and category labels, the inconsistency can be reduced via the process of edge pruning. 
% Due to the property of edge preserving, we employ the tree filter to aggregate global semantics during training.

% However, the gap still exists between low-level information and high-level semantics. To reduce this gap, we propose to construct the MST according to sparse labels.
% For a given image, we construct the graph $G$ with 4-connected neighboring edges and 4-dilated edges.
% If the two vertices of a dilated edge have the same semantic labels, we add them into $E$. Otherwise, this dilated edge will be discarded. As a result, each pixel is adjacent to up to 8 other pixels.
% After constructing G with extra label guided dilated edges, the same MST generation strategy can be utilized.
% The label guided MST has the potential to jump out of local area when aggregating semantic information.
% Given the MST of an image, similar to \cite{yang2014stereo,song2019learnable}, the distance from pixel $i$ to pixel $j$ along the hyperedge can be calculated
% by the summation of the connected edges, 
Similar to \cite{yang2014stereo,song2019learnable}, the distance between two vertices of the MST can be calculated by the weight summation of their connected edges.
% The distance of the shortest path from vertice $i$ to vertice $j$, denoted as the hyper-edge $\mathbb{E}_{i,j}$, forms the distance map of the MST, 
And the distance of the shortest path between vertices, denoted as the hyper-edge $\mathbb{E}$, forms the distance map of the MST, 
%%%%%%%%%%%%%%%%%%%%%%%%%%%%%%%%%%%%%%%%%%%%%%%%%%%%%%%
% \begin{equation}
% D^{\{low,high\}}_{i,j} = D^{\{low,high\}}_{j,i} = \sum_{(k,m)\in \mathbb{E}^{\{low,high\}}_{i,j}} \omega^{\{low,high\}}_{k,m},
% \label{eq:hyperedge_weight}
% \end{equation}
%%%%%%%%%%%%%%%%%%%%%%%%%%%%%%%%%%%%%%%%%%%%%%%%%%%%%%%
\begin{equation}
D^{*}_{i,j} = D^{*}_{j,i} = \sum_{(k,m)\in \mathbb{E}^{*}_{i,j}} \omega^{*}_{k,m},
\label{eq:hyperedge_weight}
\end{equation}
%%%%%%%%%%%%%%%%%%%%%%%%%%%%%%%%%%%%%%%%%%%%%%%%%%%%%%%
% \begin{equation}
% \begin{aligned}
% & D^{low}_{i,j} = D^{low}_{j,i} = \sum_{(k,m)\in \mathbb{E}^{low}_{i,j}} \omega^{low}_{k,m}, \\
% & D^{high}_{i,j} = D^{high}_{j,i} = \sum_{(k,m)\in \mathbb{E}^{high}_{i,j}} \omega^{high}_{k,m},
% \end{aligned}
% \label{eq:hyperedge_weight}
% \end{equation}
%%%%%%%%%%%%%%%%%%%%%%%%%%%%%%%%%%%%%%%%%%%%%%%%%%%%%%%
where $i$, $j$, $k$ and $m$ are vertice indexes, $*\in \{low, high\}$. To capture the long-range relation among vertices, we project the distance maps to positive affinity matrices,  
\begin{equation}
\begin{aligned}
% & A^{low} = \exp \left(-\frac{D^{low}}{\sigma}\right), \\
& A^{low} = \exp \left(-D^{low}/{\sigma}\right), \\
& A^{high} = \exp
\left(- D^{high}\right),
\end{aligned}
\label{eq:affinity_projection}
\end{equation}
where $\sigma$ is a preset constant value to modulate the color information.
Given a training image, the low-level affinity $A^{low}$ is static while the high-level affinity $A^{high}$ is dynamic during training. 
They capture pair-wise relations at different feature levels. 
By utilizing them jointly, complementary knowledge can be learned.

% The affinity matrices calculated on the MSTs preserve multi-level structural information of the image. 

\begin{figure}[t]
	\begin{center}
		\includegraphics[height=.75\linewidth]{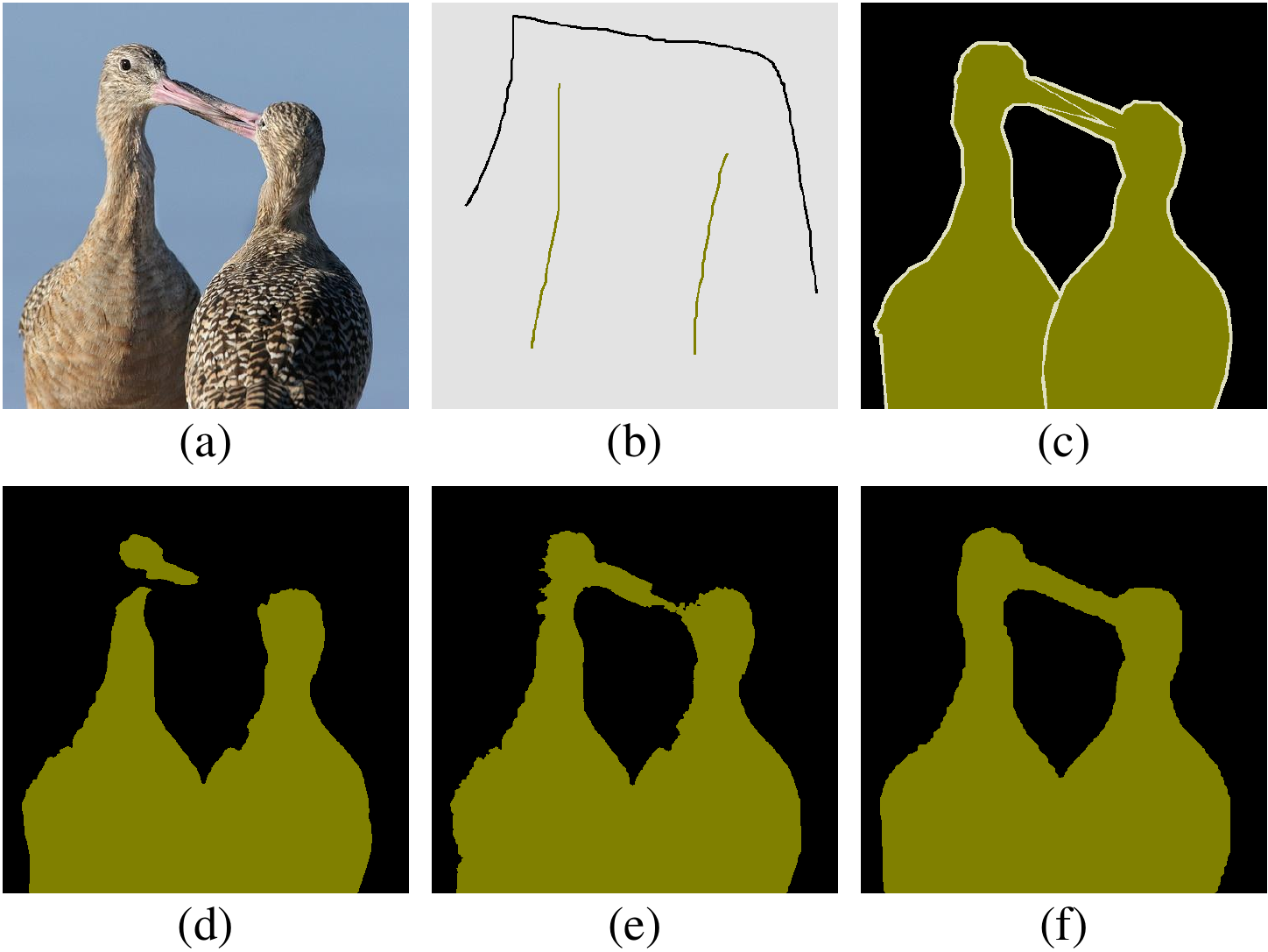}
	\end{center}
	\vspace{-0.65cm}
	\caption{Visualization of the network predictions the corresponding pseudo labels in our training framework. 
	(a)~Input image. (b)~Sparse annotation. (c)~Full annotation.
	(d)~Network prediction. (e)~Initial pseudo label generated with the low-level affinity. (f)~Final pseudo label generated with multi-level affinities.}
% 	Our method can capture the structure information of the image, which achieves higher edge accuracy in the unlabeled area. Significant differences are highlighted within the bounding boxes.}
	\vspace{-4mm}
	\label{Fig:soft_label}
\end{figure}

\noindent \textbf{Cascaded Filtering.}
Since the low-level affinity matrix $A^{low}$ contains object boundary information while the high-level affinity matrix $A^{high}$ maintains semantic consistency, we introduce a cascaded filtering strategy to generate the pseudo labels $\tilde{Y}$ from the network prediction:
% By jointly utilizing the structure information based on the low-level MST and the high-level MST, high-quality pseudo labels for unlabeled pixels can be generated. 
% Since the tree filter is a type of weighted-average filter, the category information can be aggregated in a cascaded manner:
% \begin{equation}
% \tilde{P_i}= \mathcal{F}\left[\mathcal{F}(P; A); A\right] = \frac{1}{z_i}\sum_{\forall{j}\in\Omega} A(i,j)P_j, 
% \label{eq:treeloss_aggregation}
% \end{equation}
\begin{equation}
\tilde{Y} = \mathcal{F}\left(\mathcal{F}(P , A^{low}) , A^{high}\right),
\label{eq:treeloss_aggregation}
\end{equation}
where $P$ is the prediction after the softmax operation.
By serially multiplied with low-level and high-level affinities, the network prediction can be refined in a coarse-to-fine manner, yielding high-quality soft pseudo labels. 
% The low-level and high-level structural information are aggregated in a cascading manner. 
% The low-level affinity helps preserve sharp object boundaries. However, additional noise may be introduced. Thus the high-level affinity matrix is further adopted.  
The filtering operation $\mathcal{F}(\cdot)$ is
presented as follow:
\begin{equation}
\mathcal{F}(P , A^{*}) = \frac{1}{z_i}\sum_{\forall{j}\in\Omega} A_{i,j}^{*}P_j, 
\label{eq:treeloss_filtering}
\end{equation}
where $\Omega=\Omega_{L} \cup \Omega_{U}$ is the full set of all pixels, and $z_i=\sum_{j}A_{i,j}$ is the normalization term.
To speed up the calculation of Eq.~\ref{eq:treeloss_filtering}, we adopt the efficient implementation in LTF \cite{song2019learnable} to realize the linear computational complexity.
As shown in Fig.~\ref{Fig:soft_label}, the pseudo labels generated with cascaded filtering can preserve sharper semantic boundaries than the original predictions via considering the structural information.
Since the semantic boundary is significant to semantic segmentation while mislabeled in sparse annotations, 
the performance of the segmentation model can be boosted by assigning pseudo labels for unlabeled pixels.
% The process of soft label generation can speed up by dynamic programming algorithms. We adopt the efficient implementation in LTF \cite{song2019learnable} to achieve the linear computational complexity. 

% the segmentation model can learn extra knowledge from
% the pseudo labels.
% the aggregated category information can serve as the soft label for unlabeled pixels. 

\noindent \textbf{Soft Label Assignment.}
% The aggregated category-level semantic information can be regarded as the soft pseudo-labels of the segmentation model, and the tree energy loss is proposed for soft label assignment, 
Now that we obtain the pseudo labels, the TEL is designed for soft pseudo label assignment: 
\begin{equation}
\mathcal{L}_{tree}  =  \delta({P}, \tilde{Y}),
\label{eq:general_tree_loss} 
\end{equation} 
where $\delta$ is a label assignment function, which measures the distance between predicted probability $P$ and pseudo label $\tilde{Y}$. 
Some natural choices of $\delta$ can be L1 distance, L2 distance, and so on. We empirically select L1 distance as the label assignment function. For ablations about $\delta$, please refer Sec.~\ref{Sec:ablation}.
% We conduct ablation studies on the formation of $\delta({P}, \tilde{Y})$ in Sec.~\ref{Sec:ablation}.
% We evaluated different forms of $F_{assign}$ and report the experimental results in Sec.~\ref{Sec:ablation}. 
In this way, the final formation of TEL is described as follows,
% \begin{equation}
% \mathcal{L}_{tree}  = -\frac{1}{|\Omega_U|} \sum_{\forall{i}\in \Omega_U} {P_i}^T\tilde{P_i},
% \label{eq:tree_loss} 
% \end{equation} 
\begin{equation}
\mathcal{L}_{tree}  = -\frac{1}{|\Omega_U|} \sum_{\forall{i}\in \Omega_U} |P_i - \tilde{Y_i}|.
\label{eq:tree_loss} 
\end{equation} 

% where $\Omega_U$ is the unlabeled set of images. 
Note that the TEL only focuses on the unlabeled regions since the labeled regions are learned with explicit accurate supervision.
% Note that the performance of segmentation model is dynamically improved during training, the soft label $\tilde{P_i}$ can be improved at the same time.
Instead of generating pseudo labels from the sparse annotations, our TEL generates the soft labels from network prediction. 
Thus, the data-driven model learning procedure will benefit our online self-training strategy. 

\subsection{Discussion}\label{sec:diss}
Tree filter has been applied in many vision tasks thanks to the property of structure-preserving.
% Previous works have proved the effectiveness of the tree filter. 
Previous methods apply tree filters to original images for image smoothing \cite{bao2013tree} and stereo matching \cite{yang2014stereo}, or intermediate features for feature transform \cite{song2019learnable,song2020rethinking}. Our method is inspired by these works but for a totally different purpose. We capture both low-level and high-level affinities and apply them to network predictions for soft pseudo label generation in SASS, achieving single-stage dynamic online self-training. 
To the best of our knowledge, it is the first time that the tree filter is introduced in solving the SASS problem.
% we instead apply the tree filter to network predictions to build a bridge among multi-level image information, which enables dynamic online self-training.
% between low-level structural and high-level semantic information.

\section{Experiment}

\begin{table*}[tb]
	\centering
% 	\footnotesize
	\scalebox{1.0}{
		\begin{tabular}{lcccccccc}
			\toprule
			Method &Backbone &Publication &Supervision &Multi-stage &Alt. Opt. &Extra Data & CRF &mIoU \\ 
			\midrule
			\midrule
			(1)~DeeplabV2 \cite{chen2017deeplab} &VGG16 &TPAMI'17 &Full &- &- &- &\checkmark &71.6  \\
			(2)~DeeplabV2 \cite{chen2017deeplab} &ResNet101 &TPAMI'17 &Full &- &- &- &\checkmark &77.7  \\
% 			(3)~SegSort \cite{hwang2019segsort} &ResNet101 &ICCV'19 &Full &- &- &- &- &77.3  \\
			(3)~DeepLabV3+ \cite{chen2018encoder} &ResNet101 &ECCV'18 &Full &- &- &- &- &80.2  \\
			(4)~LTF \cite{song2019learnable} &ResNet101 &NeurIPS'19 &Full &- &- &- &- &80.9  \\
			\midrule
% 			\midrule
			What's the point \cite{bearman2016s} &(1) &ECCV'16 &Point &- &- &- &- &43.4 \\
			KernelCut Loss \cite{tang2018regularized} &(2) &ECCV'18 &Point &\checkmark &- &- &\checkmark &57.0 \\ 
			A\textsuperscript{2}GNN \cite{zhang2021affinity} &(2) &TPAMI'21 &Point &\checkmark &- &- &\checkmark &66.8 \\ 
% 			\midrule
% 			{\dag}DenseCRF Loss \cite{tang2018regularized} &(3) &ECCV'18 &Point &- &- &- &61.0 \\ 
            Seminar \cite{chen2021seminar} &(3) &ICCV'21  &Point &\checkmark &- &- &- &72.5 \\
            SPML \cite{ke2021universal} &(2) &ICLR'21 &Point &- &- &\checkmark &\checkmark &\textcolor{blue}{\textit{73.2}} \\
            TEL &(3) &CVPR'22 &Point &- &- &- &- &64.9 \\
			TEL &(4) &CVPR'22 &Point &- &- &- &- &68.4 \\
			TEL w. Seminar &(3) &CVPR'22 &Point &\checkmark &- &- &- &\textcolor{red}{\textbf{74.2}}\\ 
			\midrule
% 			\midrule
			ScribbleSup \cite{lin2016scribblesup} &(1) &CVPR'16 &Scribble &\checkmark &\checkmark &- &\checkmark &63.1 \\
			NormCut Loss \cite{tang2018normalized} &(2) &CVPR'18 &Scribble &\checkmark &- &- &\checkmark &74.5 \\
			DenseCRF Loss \cite{tang2018regularized} &(2) &ECCV'18 &Scribble &\checkmark &- &- &\checkmark &75.0 \\
			KernelCut Loss \cite{tang2018regularized} &(2) &ECCV'18 &Scribble &\checkmark &- &- &\checkmark &75.0 \\
			GridCRF Loss \cite{marin2019beyond} &(2) &ICCV'19 &Scribble &\checkmark &\checkmark &- &- &72.8 \\
			BPG \cite{wang2019boundary} &(2) &IJCAI'19 &Scribble &- &- &\checkmark &- &76.0 \\
			SPML \cite{ke2021universal} &(2) &ICLR'21 &Scribble &- &- &\checkmark &\checkmark &76.1 \\
			URSS \cite{pan2021scribble} &(2) &ICCV'21 &Scribble &\checkmark &- &- &\checkmark &76.1 \\ 
			PSI \cite{xu2021scribble} &(3) &ICCV'21 &Scribble &- &\checkmark &- &- &74.9 \\ 
			Seminar \cite{chen2021seminar} &(3) &ICCV'21  &Scribble &\checkmark &- &- &- &76.2 \\
			A\textsuperscript{2}GNN \cite{zhang2021affinity} &(4) &TPAMI'21 &Scribble &\checkmark &- &- &\checkmark &76.2 \\
% 			\midrule
% 			{\dag}DenseCRF Loss \cite{tang2018regularized} &(3) &ECCV'18 &Scribble &- &- &- &75.0 \\ 
			TEL &(3) &CVPR'22  &Scribble &- &- &- &- &\textcolor{blue}{\textit{77.1}} \\
			TEL &(4) &CVPR'22  &Scribble &- &- &- &- &\textcolor{red}{\textbf{77.3}}\\
			 
			\bottomrule
	\end{tabular}}
	\vspace{-0.2cm}
	\caption{Experimental results of the point- and the scribble-annotated semantic segmentation methods on the Pascal VOC 2012 validation set. 
% 	\dag~ means the reproduced results in our training and testing settings.
    Experimental settings for multi-stage training (Multi-stage), 
    alternating optimization (Alt. Opt.),
    extra supervised data (Extra Data) during training, and CRF post-processing (CRF) during testing are considered.
% 	The proposed tree energy loss outperforms the state-of-the-art methods without multi-stage training procedure (Multi-stage), extra supervised data (Extra data) during training, or CRF post-processing (CRF) during testing.
    Top two results are highlighted in \textcolor{blue}{\textit{blue}} and \textcolor{red}{\textbf{red}}.} 
	\label{Tab:cmp_point_scibble}
\vspace{-4mm}
\end{table*}

\subsection{Datasets and Annotations}
\noindent \textbf{Datasets.} 
Pascal VOC 2012 \cite{everingham2010pascal} contains 20 object categories and a background class. Following previous methods \cite{tang2018regularized,pan2021scribble,xu2021scribble,chen2021seminar}, the augmented dataset \cite{hariharan2011semantic} with 10,582 training and 1,449 validation images are used.
Cityscapes \cite{cordts2016cityscapes} is built for urban scenes. It consists of 2,975, 500, 1,525 fine-labeled images for training, validation, and testing, respectively. There are a total of 30 annotated classes in the dataset, and 19 of which are used for semantic segmentation. 
ADE20k \cite{zhou2017scene} is a challenging benchmarks with 150 fine-grained classes. It collects 20,210, 2,000, and 3,352 images for training, validation, and testing.

\noindent \textbf{Annotations.} 
For point-supervised and scribble-supervised settings, the point-wise annotation \cite{bearman2016s} and the scribble-wise annotation \cite{lin2016scribblesup} of Pascal VOC 2012 dataset are respectively used. 
For the block-supervised setting, we synthesize the block-wise annotations on Cityscapes and the ADE20k datasets. Specially, given the full annotations, we remove the labeled pixels sequentially from semantic edges to interior regions until the ratio of the rest labeled pixels reaches the preset threshold.
% Examples of synthetic block-wise annotations are illustrated in Fig.~\ref{Fig:block_annotations}.
Examples of synthetic block-wise annotations can be found in our supplementary material.

% \begin{figure*}[t]
% 	\begin{center}
% 		\includegraphics[width=1.\linewidth]{Figs/figure4.pdf}
% 	\end{center}
% 	\vspace{-6mm}
% 	\caption{The block-wise annotations for Cityscapes dataset. 
% 	The percentage value denotes the ratio of rest annotations after label removal. Ratio 100\% is the full label of the image.}
% % 	\vspace{-2mm}
% 	\label{Fig:block_annotations}
% \end{figure*}

\subsection{Implementation Details}
We adopt three popular semantic segmentation models (i.e., the DeeplabV3+ \cite{chen2018encoder}, the LTF \cite{song2019learnable}, and the HRNet \cite{sun2019high}) for experiments. 
The ResNet-101 \cite{he2016deep} and the HRNetW48 \cite{sun2019high} pre-trained on ImageNet \cite{deng2009imagenet} dataset are used as backbone networks. 
For data augmentation, random horizontal flip, random resize in $[0.5, 2.0]$, random crop, and random brightness in $[-10, 10]$ are employed.
The input resolutions are $512 \times 512$, $1024 \times 512$, and $512 \times 512$ for Pascal VOC 2012, Cityscapes, and ADE20k datasets, respectively.
And the corresponding initial learning rates are $0.001$, $0.01$, and $0.02$.
% For Pascal VOC 2012 dataset, the $512 \times 512$ training images are fed to the model with the initial learning rate $0.001$.
% For Cityscapes dataset, the $1024 \times 512$ training images are fed to the model with the initial learning rate $0.01$.
% For ADE20k dataset, the $512 \times 512$ training images are fed to the model with the initial learning rate $0.02$.
The SGD optimizer with the momentum of $0.9$, weight decay $1e^{-4}$ polynomial schedule is utilized. 
% All the experiments for tree energy loss are carried out in a single-stage manner. 
% The model with TEL can be trained in a single-stage manner.
The total training iterations are 80k, 40k, and 150k for Pascal VOC 2012, Cityscapes, and ADE20k datasets, respectively.
In our practice, we set $\lambda=0.4$ in Eq.~\ref{eq:total_loss}. 
As for $\sigma$ in Eq.~\ref{eq:affinity_projection}, we set $\sigma=0.02$ in Pascal VOC 2012 dataset, and $\sigma=0.002$ in Cityscapes and ADE20k datasets due to the low-level appearance variety of the semantic categories.
All experiments are conducted on Pytorch \cite{paszke2019pytorch} with 4 Tesla V100 (32G) GPUs.

\subsection{Comparison with State-of-the-art Methods}
\noindent \textbf{Point-wise supervision.} 
Point-supervised semantic segmentation is an extreme setting in SASS. What's the Point \cite{bearman2016s} provides the point-wise annotations for Pascal VOC 2012 dataset. However, it only labels the foreground classes and lacks the annotations of the background class. Following previous works \cite{obukhov2019gated, chen2021seminar}, we adopt the point-wise background annotations from the scribble annotations in ScribbleSup \cite{lin2016scribblesup}. 
% Since the DeeplabV2 \cite{chen2017deeplab} is pre-trained on the fully annotated COCO dataset, which is against the purpose of the SASS task, we choose DeeplabV3+ \cite{chen2018encoder} and LTF \cite{song2019learnable} as the basic segmentation models instead.
% As shown in Table~\ref{Tab:cmp_point_scibble}, 
% the reproduced DenseCRF Loss trained in a single-stage manner achieves 58.5\% mIoU while the proposed tree energy loss achieves 64.9\% mIoU. 
% The DeeplabV3+ \cite{chen2018encoder} and LTF \cite{song2019learnable} are selected as basic segmentation models and 
The experimental results are reported in Tab.~\ref{Tab:cmp_point_scibble}.
% Among the single-stage training methods, the proposed tree energy loss achieves significant performance improvement. 
% When equipped with LTF \cite{song2019learnable}, our method without CRF post-processing achieves 1.6\% mIoU higher than the two-stage method A\textsuperscript{2}GNN.
When equipped with DeeplabV3+, our baseline employing the partial cross-entropy loss can produce 58.5\% mIoU. Incorporating the TEL, the segmentation model achieves 6.4\% mIoU improvements compared with our baseline.
% our method can achieve 64.9\% mIoU with a single-stage training strategy, \textcolor{blue}{which is 6.4\% higher than our baseline model.}
% When equipped with Deeplabv3+, our single-stage baseline model just using the partial cross-entropy loss achieves 58.5 mIoU. Our method achieves 6.4\% improvement by incorporating the TLE.
It demonstrates that our method is effective and easy to be plugged into the existing segmentation frameworks.
% Among recent methods, Seminar \cite{chen2021seminar} achieves the best performance. 
Among recent methods, Seminar \cite{chen2021seminar} has a similar workflow with the semi-supervised mean-teacher method \cite{tarvainen2017mean} and achieves 72.5\% mIoU.
% However, the total hyper-parameters and the model volume increase due to the complex training procedure.
% Thanks to the simplicity of our method, 
We apply our method to the Seminar by replacing the DenseCRF loss with the proposed TEL. It shows that TEL can bring additional 1.7\% mIoU improvements and achieve state-of-the-art performance.

\noindent \textbf{Scribble-wise supervision.} 
% We compare our tree energy loss with state-of-the-art methods including ScribbleSup \cite{lin2016scribblesup}, NormCut Loss \cite{tang2018normalized}, DenseCRF Loss \cite{tang2018regularized}, KernelCut Loss \cite{tang2018regularized}, GridCRF Loss \cite{marin2019beyond}, BPG \cite{wang2019boundary}, SPML \cite{ke2021universal} and A\textsuperscript{2}GNN \cite{zhang2021affinity}. 
As shown in Tab.~\ref{Tab:cmp_point_scibble}, the proposed TEL can be applied in the single-stage training framework and calls for no additional supervised data during training or CRF post-processing during testing.
ScribbleSup \cite{lin2016scribblesup} presents an alternative proposal generation and model training method and achieves 63.1\% mIoU. 
To achieve higher performance, the regularized losses are designed by mining pair-wise relations from low-level image information. BPG \cite{wang2019boundary} and SPML \cite{ke2021universal} utilize the edge detectors (i.e., the pre-trained HED method \cite{xie2015holistically}) for semantic edge generation and over-segmentation. However, extra supervised data are required to learn the edge detectors.
Among all the recent methods, A\textsuperscript{2}GNN achieves the best performance. It first generates seed labels by mixing up multi-level supervisions, then refines the seed labels with affinity attention graph neural networks. Finally, the CRF post-processing is adopted.
Compared with A\textsuperscript{2}GNN, our method can be trained in a single-stage manner while outperforming it by 1.1\% mIoU without any post-processing.

\begin{figure}[t]
	\begin{center}
		\includegraphics[width=1.\linewidth]{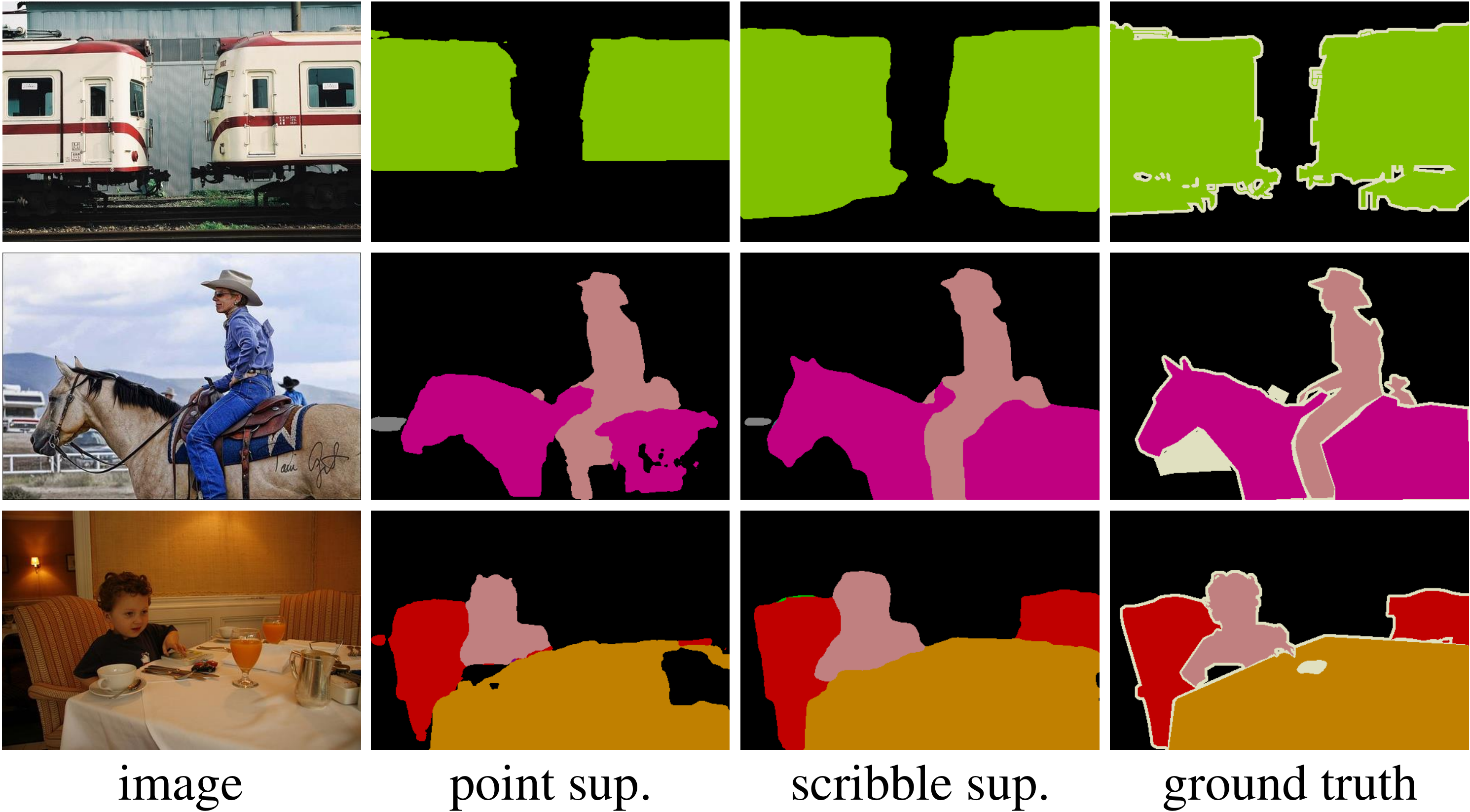}
	\end{center}
	\vspace{-6mm}
	\caption{Qualitative results for the proposed TEL on Pascal VOC 2012 dataset. The point sup. and the scribble sup. indicate the point and the scribble supervision, respectively.}
	\vspace{-2mm}
	\label{Fig:point_scribble_vis}
\end{figure}

Fig.~\ref{Fig:point_scribble_vis} illustrates some qualitative results on Pascal VOC 2012 dataset.
Although the annotations are quite sparse, our method can leverage the structure information among labeled and unlabeled regions and generate promising masks with fine semantic boundaries.

\noindent \textbf{Block-wise supervision.} 
To further evaluate the robustness of TEL, we carry out additional experiments with block-wise annotations. 
Notice that the Pascal VOC 2012 dataset is relatively easy since the prediction of pixels close to the semantic boundaries is usually ignored in accuracy calculation (shown in Fig.~\ref{Fig:annotations}(b)), so we resort to the Cityscapes and the ADE20k datasets.
% for the block-wise SASS experiments.
To evaluate the performance with different sparsity, we generate the block-wise annotations at three different levels, including 10\%, 20\%, and 50\% of full labels. 
The 100\% ratio indicates the fully annotated setting, which servers as the upper bound of the SASS methods. 
The baseline is the segmentation network trained with the partial cross-entropy loss only. 
We compare our TEL with the state-of-the-art DenseCRF Loss \cite{tang2018regularized} and report the results in Tab.~\ref{Tab:cmp_block}.
For all block-annotated settings, we use the default hyper-parameters for DenseCRF Loss reported in the paper and it achieves higher accuracy compared with the baseline. 
% However, the performance improvement is limited in the complex scenes such as dozens of appearing objects in Cityscapes and hundreds of semantic categories in ADE20k. 
However, the performance improvement is relatively limited. 
The proposed TEL captures both the low-level and the high-level relation and outperforms the DenseCRF Loss in all block-supervised settings. 

\subsection{Ablation Study}\label{Sec:ablation}
% We carried out thorough ablation studies on the block-supervised SASS setting, where the ratio of labeled pixels is 20\% in the Cityscapes dataset. The HRNet is selected as our basic segmentation network. The single-scale testing results are reported in Table~\ref{Tab:abl_total}. 
% We carried out thorough ablation studies on the scribble-supervised setting of the Pascal VOC 2012 dataset. The DeeplabV3+ is selected as our basic segmentation network. 
% The total quantification results are reported in Tab.~\ref{Tab:abl_total}. 
We perform thorough ablation studies for TEL. The scribble-supervised results of DeeplabV3+ on the Pascal VOC 2012 dataset are reported unless otherwise stated.

\noindent \textbf{Loss formation.}
% The proposed tree energy loss generates soft labels for unlabeled pixels based on both low-level and high-level structural information. 
TEL learns to assign soft labels for unlabeled pixels.
The experiments about the loss formation in Eq.~\ref{eq:general_tree_loss} are carried out to evaluate the effectiveness of the TEL. 
The baseline model achieves 68.8\% with partial cross-entropy loss.
% As shown in Table~\ref{Tab:abl_lossForm}, the first row means baseline model without any formation of tree energy loss. It achieves 58.6\% mIoU.
As shown in Tab.~\ref{Tab:abl_lossForm}, the performance can be improved by different formations of TEL. Among them, the L1 distance achieves the best result with 77.1\% mIoU, thus we choose it as the final implementation of our TEL.

\begin{table*}[tb]
	\centering
% 	\footnotesize
	\scalebox{0.9}{
		\begin{tabular}{c|c|c|cccc|cccc}
			\toprule
			\multirow{2}{*}{Model} &\multirow{2}{*}{Backbone}  &\multirow{2}{*}{Method} & \multicolumn{4}{c|}{Cityscapes} & \multicolumn{4}{c}{ADE20k} \\
% 			\cmidrule(r){4-7}
			&& &10\% &20\% &50\% &100\% &10\% &20\% &50\% &100\% \\ 
			\midrule
			\midrule
			\multirow{3}{*}{HRNet} &\multirow{3}{*}{HRNetW48} &Baseline &52.8 &58.6 &68.8 &78.2 &30.2 &33.1 &37.2 &42.5  \\
			                       &&DenseCRF Loss \cite{tang2018regularized} &57.4 &61.8 &70.9 &- &31.9 &33.8 &38.4 &- \\
			                       &&TEL &\textcolor{black}{\textbf{61.9}} &\textcolor{black}{\textbf{66.9}} &\textcolor{black}{\textbf{72.2}} &- &\textcolor{black}{\textbf{33.8}} &\textcolor{black}{\textbf{35.5}} &\textcolor{black}{\textbf{40.0}} &-  \\ 
			\midrule
			\multirow{3}{*}{DeeplabV3+} & \multirow{3}{*}{ResNet101} &Baseline &48.4 &52.8 &60.5 &80.2 &30.8 &33.4 &36.6 &44.6  \\ 
			                       &&DenseCRF Loss \cite{tang2018regularized} &55.6 &61.5 &69.3 &- &31.2 &34.0 &37.4&-  \\
			                       &&TEL &\textcolor{black}{\textbf{64.8}} &\textcolor{black}{\textbf{67.3}} &\textcolor{black}{\textbf{71.5}} &- &\textcolor{black}{\textbf{34.3}} &\textcolor{black}{\textbf{36.0}} &\textcolor{black}{\textbf{39.2}} &-  \\ 
			\bottomrule
	\end{tabular}}
	\vspace{-0.2cm}
	\caption{Single-stage training results for block-wise annotations on Cityscapes and ADE20k validation sets.} 
	\label{Tab:cmp_block}
% \vspace{-2mm}
\end{table*}

\begin{table*}[ht]
\center
\begin{minipage}{0.3\linewidth}
    \centering
	\resizebox{\columnwidth}{!}{
	\setlength\tabcolsep{3.5pt}
	\footnotesize
		\begin{tabular}{l|c|c}
			\toprule
			Formation & Equation & mIoU \\
			\midrule
			Cross Entropy & $-\sum{\tilde{Y}\log{P}} $ &76.0 \\
			Dot Product &$-\sum{P^T\tilde{Y}}$  &76.6 \\
			L2 Distance & $\sum{{|P-\tilde{Y}|}^2}$ &75.1 \\
			L1 Distance  & $\sum{|P-\tilde{Y}|}$ &\textbf{77.1} \\
			\bottomrule
	\end{tabular}}
	\subcaption{\footnotesize{Ablation of the loss formation for TEL in Eq.~\ref{eq:general_tree_loss}.}}
	\label{Tab:abl_lossForm}
\end{minipage} %\par
\hspace{2mm}
\begin{minipage}{0.3\linewidth}
    \centering
	\resizebox{\columnwidth}{!}{
	\setlength\tabcolsep{3.25pt}
	\footnotesize
		\begin{tabular}{cc|c}
			\toprule
			Low-level        &High-level        &mIoU \\
			\midrule
			                &               &68.8 \\
			\checkmark      &               &76.3 \textcolor{cyan}{{(+7.5)}}\\
			                &\checkmark     &71.9 \textcolor{cyan}{{(+3.1)}} \\
			\checkmark      &\checkmark     &\textbf{77.1} \textcolor{cyan}{{\textbf(+8.3)}} \\
			\bottomrule
	\end{tabular}}
% 	\subcaption{\footnotesize{Effect of the components for TEL in terms of the low-level and high-level affinities.}}
    \subcaption{\footnotesize{Effect of the the low-level and the high-level affinities in TEL.}}
	\label{Tab:abl_module}
\end{minipage} %\par
\hspace{2mm}
\begin{minipage}{0.3\linewidth}
    \centering
	\resizebox{\columnwidth}{!}{
	\setlength\tabcolsep{6pt}
    \footnotesize
		\begin{tabular}{c|c|c}
			\toprule
			Information                 &Method     &mIoU \\
			\midrule
			\multirow{2}{*}{Low-level}  &BF         &75.0 \\
			                            &MST        &\textbf{76.3} \textcolor{cyan}{\textbf{(+1.3)}}\\
			\midrule                            
			\multirow{2}{*}{High-level} &NL         &70.2 \\
			                            &MST        &\textbf{71.9} \textcolor{cyan}{\textbf{(+1.7)}}\\                           
			\bottomrule
	\end{tabular}}
% 	\vspace{2mm}
	\subcaption{\footnotesize{Impact of affinity generation methods based on different levels of image information.}}
	\label{Tab:abl_kernel}
\end{minipage} %\par
\vspace{2mm}

\begin{minipage}{0.3\linewidth}
\centering
	\resizebox{\columnwidth}{!}{
	\setlength\tabcolsep{6pt}
	    \footnotesize
		\begin{tabular}{c|ccc}
			\toprule
			Variant &LH-P &HL-C &LH-C \\
			\midrule
			mIoU &76.4 &75.8 &\textbf{77.1}  \\
			\bottomrule
	\end{tabular}}
	\subcaption{\footnotesize{Affect of the affinity aggregation strategies.}}
	\label{Tab:abl_aggregation}
\end{minipage} %\par
\hspace{2mm}
\begin{minipage}{0.3\linewidth}
    \centering
	\resizebox{\columnwidth}{!}{
	\setlength\tabcolsep{3.5pt}
		\footnotesize
		\begin{tabular}{c|ccccc}
			\toprule
			$\lambda$ & 0.1 &0.2 &0.3 &0.4 &0.5 \\
			\midrule
			mIoU &74.9 &76.0 &76.4 &\textbf{77.1} &77.0 \\
			\bottomrule
	\end{tabular}} 
	\subcaption{\footnotesize{Effectiveness evaluation of $\lambda$ in Eq.~\ref{eq:total_loss}.}}
	\label{Tab:abl_hp_lambda}
\end{minipage} %\par
\hspace{2mm}
\begin{minipage}{0.3\linewidth}
    \centering
	\resizebox{\columnwidth}{!}{
	\setlength\tabcolsep{3pt}
		\footnotesize
		\begin{tabular}{c|ccccc}
			\toprule
			$\sigma$ & 0.01 &0.02 &0.03 &0.04 &0.05 \\
			\midrule
			mIoU &76.6 &\textbf{77.1} &76.8 &77.0 &76.3 \\
			\bottomrule
	\end{tabular}}
	\subcaption{\footnotesize{Effectiveness evaluation of $\sigma$ in Eq.~\ref{eq:affinity_projection}.}}
	\label{Tab:abl_hp_sigma}
\end{minipage} %\par
\vspace{-2mm}
\caption{Ablation studies on the proposed TEL. We train on the scribble annotations and test on the Pascal VOC 2012 validation set.}\label{Tab:abl_total}
\vspace{-4mm}
\end{table*}

\noindent \textbf{Affinity level.}
The TEL leverage both low-level and high-level structural information to generate pseudo labels for unlabeled pixels. 
To evaluate their effectiveness, we carry out ablation studies in Tab.~\ref{Tab:abl_module}. Compared with the baseline, introducing low-level and high-level information can achieve 7.5\% and 3.1\% mIoU improvement, respectively.
Adopting both of them, our method achieves 77.1\% mIoU, which is 8.3\% higher than the baseline. 

% \begin{table}[tb]
% 	\centering
% 	\footnotesize
% 	\resizebox{\columnwidth}{!}{
% % 	\setlength\tabcolsep{6pt}
% 		\begin{tabular}{c|c|c}
% 			\toprule
% 			Formation & Equation & mIoU \\
% 			\midrule
% % 			\midrule
% 			- & - &58.6 \\
% 			Cross Entropy & $-\sum{\tilde{Y}\log{p}} $ &65.3 \\
% 			Dot Product &$-\sum{p^T\tilde{Y}}$  &66.1 \\
% 			MSE Loss & $\sum{{|p-\tilde{Y}|}^2}$ &63.7 \\
% 			L1 Loss  & $\sum{|p-\tilde{Y}|}$ &\textbf{66.9} \\
% 			\bottomrule
% 	\end{tabular}}
% % 	\vspace{-0.2cm}
% 	\caption{Analysis about the formation of the tree energy loss in Eq.~\ref{eq:general_tree_loss} on Cityscapes validation set.}
% 	\label{Tab:abl_lossForm}
% \vspace{-6mm}
% \end{table}

\begin{figure}[t]
	\begin{center}
		\includegraphics[width=0.95\linewidth]{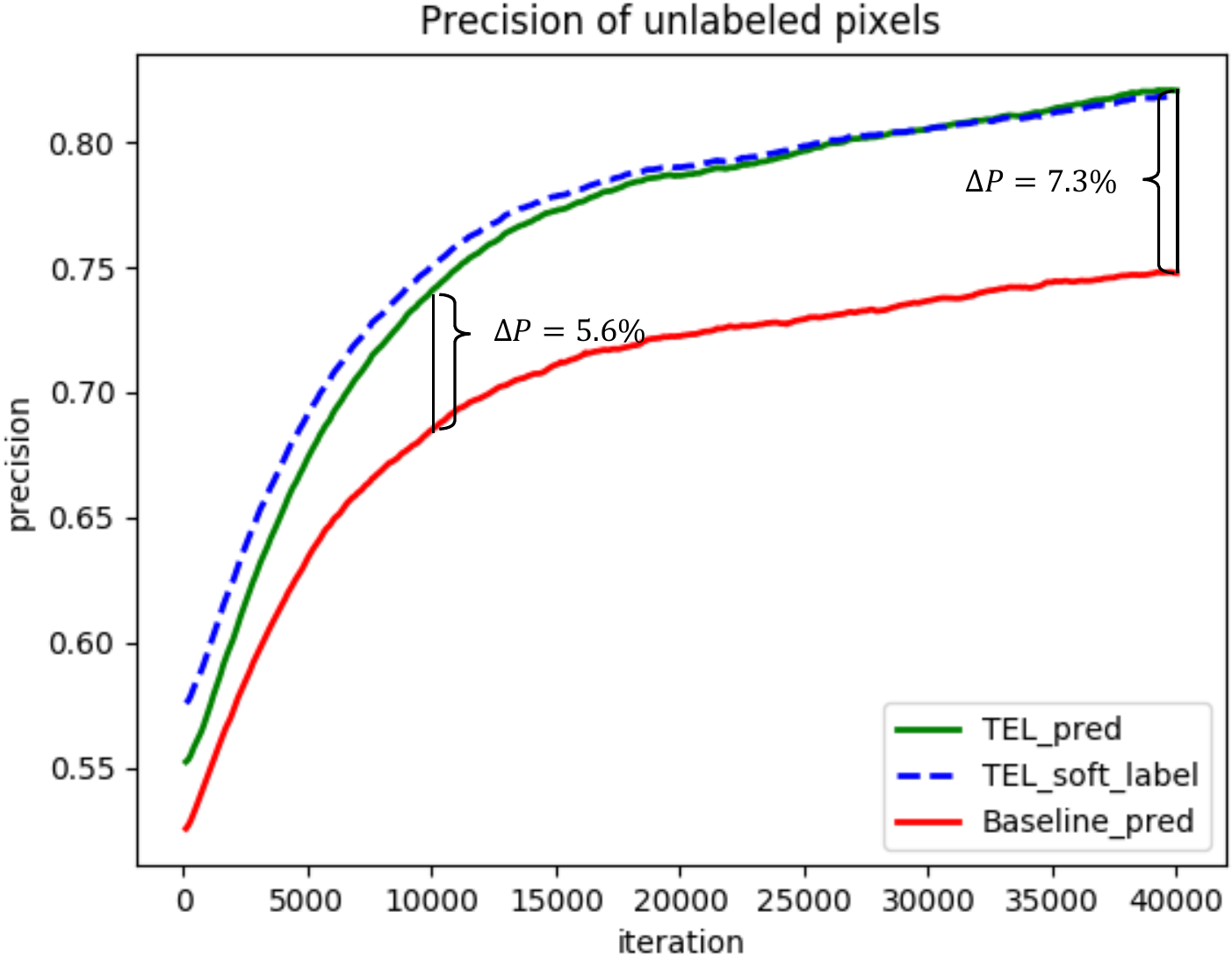}
	\end{center}
	\vspace{-6mm}
	\caption{Qualitative evaluation about the network predictions and the soft pseudo labels in unlabeled regions during training. The results of 20\% block-wise annotations on Cityscapes are illustrated. 
	`TEL\_pred' and `TEL\_soft\_label' are the network predictions and the generated pseudo labels of the TEL model, respectively.
	${\Delta}P$ denotes the precision difference between the baseline training framework and the proposed TEL framework. Extra knowledge can be learned by the segmentation network by incorporating TEL into the baseline.
	}
	\vspace{-4mm}
	\label{Fig:abl_pred_prec}
\end{figure}

\noindent \textbf{Affinity generation.}
The TEL captures the low-level and high-level structural information to generate the affinity matrices in Eq.~\ref{eq:affinity_projection}. 
As shown in Tab.~\ref{Tab:abl_kernel}, we compare different methods of pair-wise affinity generation, including the Bilateral Filter (BF) for low-level affinity and the Non-Local operation (NL) for high-level affinity.
The implementations for BF \cite{tang2018regularized} and NL \cite{wang2018non} are adopted. 
% The BF and lowMST leverage the low-level image information to calculate the pair-wise affinity. 
Our method generates affinity matrices based on the MST.
Compared with BF, our method requires fewer hyper-parameters while achieving higher accuracy. 
% which demonstrates the effectiveness of low-level affinity generation.
As for high-level affinity, our method achieves 1.7\% higher mIoU than NL.
These results demonstrate the effectiveness of TEL in both the low-level and the high-level affinity generation.
% For the utilization of the high-level semantic information, the highMST outperforms the NL operation, which demonstrates the highMST can model high-level structure information effectively.
% Our complete method uses both the lowMST and the highMST to capture the multi-level structural information, achieving the best performance. 

% \textbf{Aggregation strategy for multi-level information.}
\noindent \textbf{Affinity aggregation.}
How to aggregate the multi-level information is significant to pseudo-label generation. We construct different variants of TEL based on the aggregation strategy. 
As shown in Tab.~\ref{Tab:abl_aggregation}, LH-P denotes the variant of parallel aggregation. 
In this case, the low-level and the high-level affinity matrices are multiplied with network predictions separately to produce two pseudo labels, and both of them are used as the guidance for unlabeled pixels. 
In contrast to the parallel aggregation strategy, the cascading aggregation strategies merge network predictions with the multi-level affinity matrices one by one to refine the pseudo labels sequentially.
% Among these aggregation strategies, we find that aggregating the low-level information first (denoted as LH-C) achieve the best result.
Among the cascading strategies, we find that aggregating the low-level information first (denoted as LH-C) achieves better results than the variant of aggregating the high-level information first (denoted as HL-C).
The low-level affinity is generated from the static color information, which may bring noise due to the inconsistency between low-level color and high-level category information. Incorporating the learnable high-level affinity can improve the semantic consistency. 
% \textcolor{red}{The low-level affinity contains the intrinsic structural information of an image, which provides initial priors for pseudo label generation. However, it may bring noise due to the inconsistency between low-level vision and high-level category information. To alleviate this issue, the high-level affinity is further aggregated.}
% Thus the cascading aggregation strategy of LH-C is adopted and it achieves the best result on Pascal VOC 2012 dataset. 
% We adopts the cascading aggregation strategy of LH-C and achieves 77.1\% mIoU on Pascal VOC 2012 dataset. 

\noindent \textbf{Hyper-parameters.}
We evaluate the hyper-parameters of our method, including
the $\lambda$ in Eq.~\ref{eq:total_loss} and the $\sigma$ in Eq.~\ref{eq:affinity_projection}. 
$\lambda$ is the factor to balance the segmentation loss and TEL. 
The results are reported in Tab.~\ref{Tab:abl_hp_lambda}, and we choose $\lambda$ = 0.4 for our TEL.
$\sigma$ is a normalization term for the low-level affinity matrix projection.
We evaluate the influence of $\sigma$ and report the results in Tab.~\ref{Tab:abl_hp_sigma}. 
The value of $\sigma$ is not sensitive to the segmentation accuracy and the highest mIoU is obtained when $\sigma$ equals 0.02 on the Pascal VOC 2012 dataset.
% We evaluate the hyper-parameters of our method, including the $\sigma$ in Eq.~\ref{eq:affinity_projection} and the $\lambda$ in Eq.~\ref{eq:total_loss}. 
% $\sigma$ is a normalization term for the low-level affinity matrix projection. 
% We evaluate the influence of $\sigma$ and report the results in Tab.~\ref{Tab:abl_hp_sigma}. 
% The value of $\sigma$ is not that sensitive to the segmentation accuracy and the highest mIoU can be obtained when $\sigma$ equals 0.02 on the Pascal VOC 2012 dataset. 
% The $\lambda$ is the factor to balance the loss weight between the labeled and the unlabeled data. 
% The results are reported in Tab.~\ref{Tab:abl_hp_lambda}, and we choose $\lambda$ = 0.4 for our tree energy loss.

\noindent \textbf{Quality of pseudo labels.}
We evaluate the quality of pseudo labels for unlabeled pixels on Cityscapes dataset. The baseline segmentation model is HRNet.  As shown in Fig.~\ref{Fig:abl_pred_prec}, for the model learned with TEL, the precision of the pseudo label is higher than the network prediction at the beginning of the training process, which provides import guidance for model learning. As the number of iterations increases, the precision gap between the prediction and pseudo label is gradually narrowed while the performance of both is improved all the time. Compared with the baseline, TEL can help the segmentation model learn extra knowledge from unlabeled data and achieve performance improvement (from 5.6\% to 7.3\% mIoU during training).

\subsection{Limitations}
% TEL utilizes the low-level and high-level affinities to generate the soft pseudo labels. 
% In our method,
% the initial soft labels are generated based on the static low-level structural information, which cannot benefit from the data-driven training procedure.
% The final soft labels can be dynamically refined by introducing the high-level structural information. However, TEL ignores the inherent relation between the final soft labels and the sparse ground truth. 
% This paper aims to design a simple yet effective solution for SASS, 
% which avoids complex optimization strategies and is easy to be applied in existing segmentation frameworks.
% The potential improvement for TEL will be studied in the feature.
This paper provides a simple yet effective solution for SASS and achieves state-of-the-art performance. However, it also has some limitations. First, the low-level affinity is generated from the static image, which may bring about noise in the pseudo label. For example, objects with different categories may have similar color information. Second, TEL ignores the inherent relation between the pseudo label and the sparse ground truth. Learning from noise label \cite{han2018co} and alternative optimizer like \cite{marin2019beyond} are possible solutions to solve these problems, respectively.
% This paper aims to design a simple yet effective solution for SASS.
% The potential improvement for TEL will be studied in the future.

% can train the segmentation model in a single-stage manner. 
% Our method utilizes low-level and high-level MSTs to capture the pair-wise relation among pixels. However, these MSTs are built on a single image, which cannot capture the inter-image pair-wise relation. Besides, the proposed tree energy loss generates pseudo labels from network predictions, which ignores the information from sparse annotations. The aim of our method is to propose an unified simple and effective solution for SASS. These issues will be studied in the feature.

\section{Conclusion}
This paper presents a novel tree energy loss (TEL) for sparsely annotated semantic segmentation.
The TEL captures both the low-level and the high-level structural information via minimum spanning trees to generate soft pseudo labels for unlabeled pixels, then performs online self-training dynamically. 
The TEL is effective and easy to be plugged into most of the existing semantic segmentation frameworks.
% By combining the TEL with a traditional segmentation loss, 
% an end-to-end single-stage training framework is constructed 
% This paper presents an end-to-end single-stage training framework for sparsely annotated semantic segmentation (SASS) by combining the traditional segmentation loss for labeled data and a novel tree energy loss for unlabeled ones.  
% The proposed tree energy loss captures both the low-level and the high-level structural information of the image via minimum spanning trees to generate soft pseudo labels implicitly, then performs online self-training dynamically.
Equipped with the recent segmentation model, our method can
be learned in a single-stage manner and outperforms the state-of-the-art methods in point-, scribble-, and block-wise annotated settings without alternating optimization procedures, extra supervised data, or time-consuming post-processes.

% \clearpage
%%%%%%%%% REFERENCES
{\small
\bibliographystyle{ieee_fullname}
\bibliography{egbib}
}

\clearpage

\appendix

\section{Failure cases}
\begin{figure}
	\begin{center}
		\includegraphics[width=1.\linewidth]{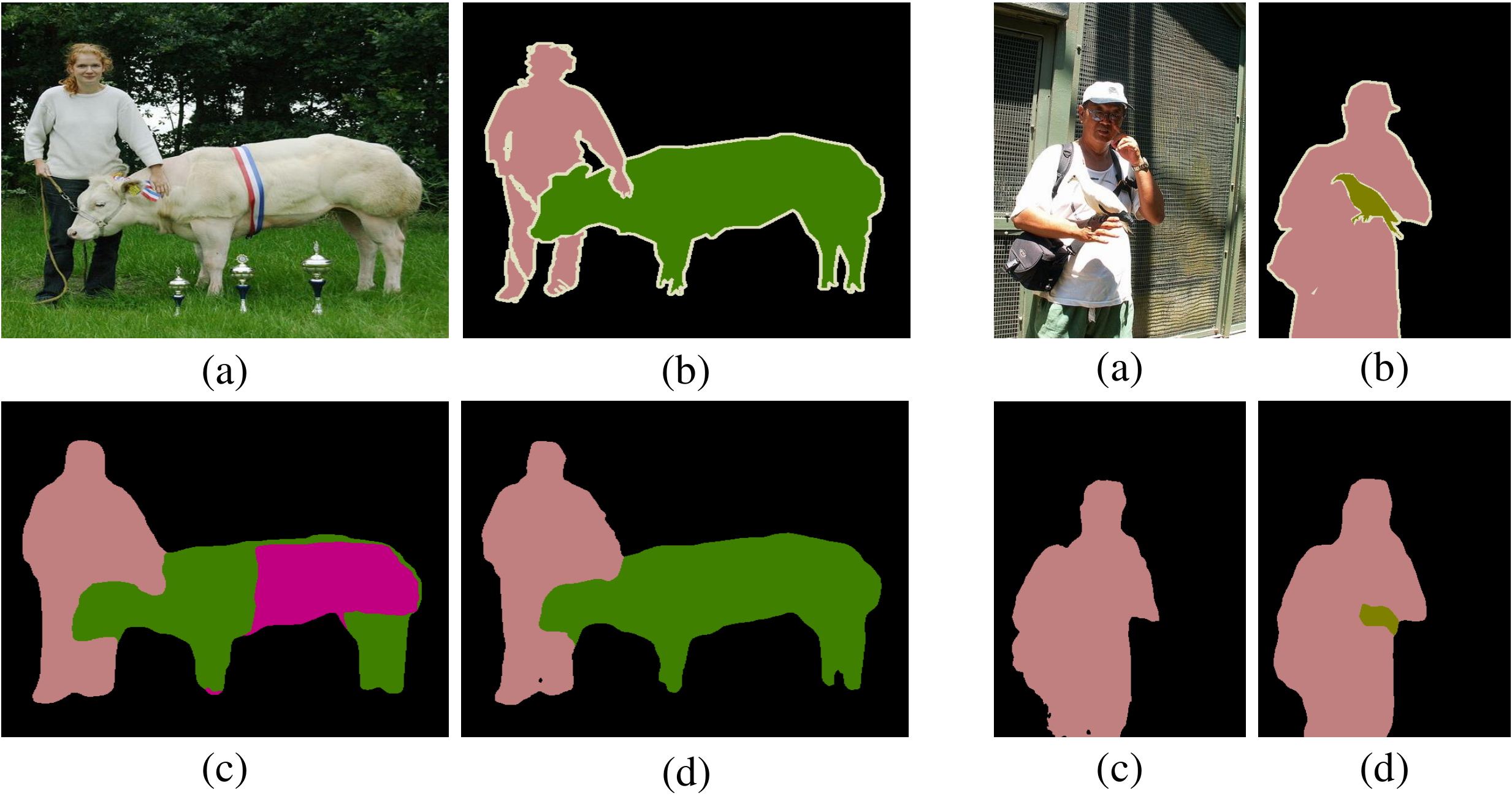}
	\end{center}
	\vspace{-6mm}
	\caption{Failure cases of the proposed method on Pascal VOC 2012  dataset \cite{everingham2010pascal}. (a)~input image. (b)~ground truth. (c)~result with point-wise supervision. (d)~result with scribble-wise supervision.}
	\vspace{-2mm}
	\label{Fig:failure_caes}
\end{figure}

Fig.~\ref{Fig:failure_caes} shows some failure cases of our method. The predictive error occurs due to the color mutation within one object region or the color similarity between two semantic objects. 
Since our approach partially relies on the low-level color prior to generate the pseudo labels, so it may make wrong predictions under point-wise supervision. 
% Fortunately, the predictive error can be reduced by introducing larger amounts of annotation during training.   
Fortunately, the predictive error can be reduced by introducing more complete annotations (i.e., scribble annotation).

\section{Computational costs}
The proposed method can be plugged into most existing semantic segmentation frameworks. During inference, the auxiliary branch can be removed. 
During training, the GPU loads and the training time evaluation on Pascal VOC 2012 dataset are reported in Tab.~\ref{Tab:gpu_load} and Tab.~\ref{Tab:time_load}, respectively. The DeeplabV3+ \cite{chen2018encoder} and the LTF \cite{song2019learnable} are selected as the baselines.
The resolution of the input image is $512 \times 512$ and the batch size is 16.
The two MST affinities are derived on the resolution
of network predictions (e.g., 4 times smaller than the
input resolution).
Leveraging the low-level information almost introduces no extra memory cost while introducing both low-level and high-level information requires 5.88\% and 3.82\% GB GPU loads for DeeplabV3+ and LTF models, respectively.
The training time of our method only increases around 9\% compared to the baseline.

\section{Comparison to the naive online self-training}
The proposed TEL leverages the multi-level structural information to dynamically generated the pseudo labels for unlabeled pixels, achieving online self-training at the training stage. As shown in Tab.~\ref{Tab:naive_online_self_train}, compared to the naive online self-training strategy which adopts a preset threshold to binary the pseudo labels, our TEL surpasses it with large improvements.

\section{Experiments on Pascal VOC 2012 test set}
The scribble-supervised experimental results on Pascal VOC 2012 test set are reported in Tab.~\ref{Tab:voc_test}.
Since the quality of pseudo labels is based on network predictions in our method, advanced backbones are helpful to fully explore the effectiveness of TEL. 
To evaluate the performance with simpler models, we also report the results of DeeplabV2 backbone.
Our TEL is clean and outperforms the multi-stage method A\textsuperscript{2}GN by more than 1\% mIoU as well.

\begin{table}[tb]
	\centering
	\setlength\tabcolsep{3pt}
% 	\footnotesize
	\scalebox{1.0}{
		\begin{tabular}{cc|cc}
			\toprule
			Low-level       &High-level     &DeeplabV3+     &LTF   \\
			\midrule
			                &               &32.85          &46.65 \\
			\checkmark      &               &32.86 \textcolor[RGB]{118,118,118}{(+0.03\%)}  &46.67 \textcolor[RGB]{118,118,118}{(+0.04\%)} \\
			\checkmark      &\checkmark     &34.78 \textcolor[RGB]{118,118,118}{(+5.88\%)}  &48.43 \textcolor[RGB]{118,118,118}{(+3.82\%)} \\
			\bottomrule
    	\end{tabular}}
    \vspace{-2mm}
    \caption{The memory cost (GB) during training, leveraging various levels of structural information.}
    \label{Tab:gpu_load}
    \vspace{-4mm}
\end{table}

\begin{table}[tb]
	\centering
	\setlength\tabcolsep{6pt}
% 	\footnotesize
	\scalebox{1.0}{
		\begin{tabular}{cc|cc}
			\toprule
			Low-level       &High-level     &DeeplabV3+     &LTF   \\
			\midrule
			                &               &9.0            &11.7 \\
			\checkmark      &               &9.2 \textcolor[RGB]{118,118,118}{(+2.2\%)}   &12.2 \textcolor[RGB]{118,118,118}{(+4.3\%)} \\
			\checkmark      &\checkmark     &9.8 \textcolor[RGB]{118,118,118}{(+8.9\%)}   &12.7 \textcolor[RGB]{118,118,118}{(+8.5\%)} \\
			\bottomrule
    	\end{tabular}}
    \vspace{-2mm}
    \caption{The training time (hour) evaluation of our method, leveraging various levels of structural information.}
    \label{Tab:time_load}
    \vspace{-4mm}
\end{table}

\begin{table}[tb]
	\centering
	\setlength\tabcolsep{5pt}
	\scalebox{1.0}{
		\begin{tabular}{c|cccccc|c}
			\toprule
			Threshold       &0.5    &0.6    &0.7    &0.8    &0.9    &0.95 &1.0 \\
			\midrule
			mIoU            &7.3    &7.3    &7.5    &56.5   &70.2   &\textbf{70.3} &68.8\\
			\bottomrule
    	\end{tabular}}
    \vspace{-2mm}
    \caption{The scribble-annotated experiments about naive online self-training on VOC 2012 val set. 
    The threshold 1.0 means the original baseline without pseudo-label supervision.
    }
    \label{Tab:naive_online_self_train}
    \vspace{-4mm}
\end{table}

\begin{table}[tb]
	\centering
% 	\footnotesize
	\scalebox{0.7}{
		\begin{tabular}{llcccc}
			\toprule
			Method &Backbone &Multi-stage &Extra Data &CRF &mIoU \\ 
			\midrule
			\midrule
			SPML\quad(ICLR'21) &DeeplabV2 &- &\checkmark &\checkmark &76.4 \\
			A\textsuperscript{2}GNN (TPAMI'21) &DeeplabV2 &\checkmark &- &\checkmark &74.0 \\
			A\textsuperscript{2}GNN (TPAMI'21) &LTF &\checkmark &- &\checkmark &76.1 \\
			TEL &DeeplabV2 &- &- &- & 74.8\\
			TEL &DeeplabV2 &- &- &\checkmark & 76.0\\
% 			TEL &DeeplabV3+ &-  &- &- &77.2\\
			TEL &LTF &-  &- &- &77.5\\
			\bottomrule
	\end{tabular}}
	\vspace{-0.2cm}
	\caption{Experimental results of the scribble-annotated semantic segmentation methods on the Pascal VOC 2012 test set.} 
	\label{Tab:voc_test}
\vspace{-4mm}
\end{table}

\section{Visualizations of block-supervised settings}
\begin{figure*}[t]
	\begin{center}
		\includegraphics[width=1.\linewidth]{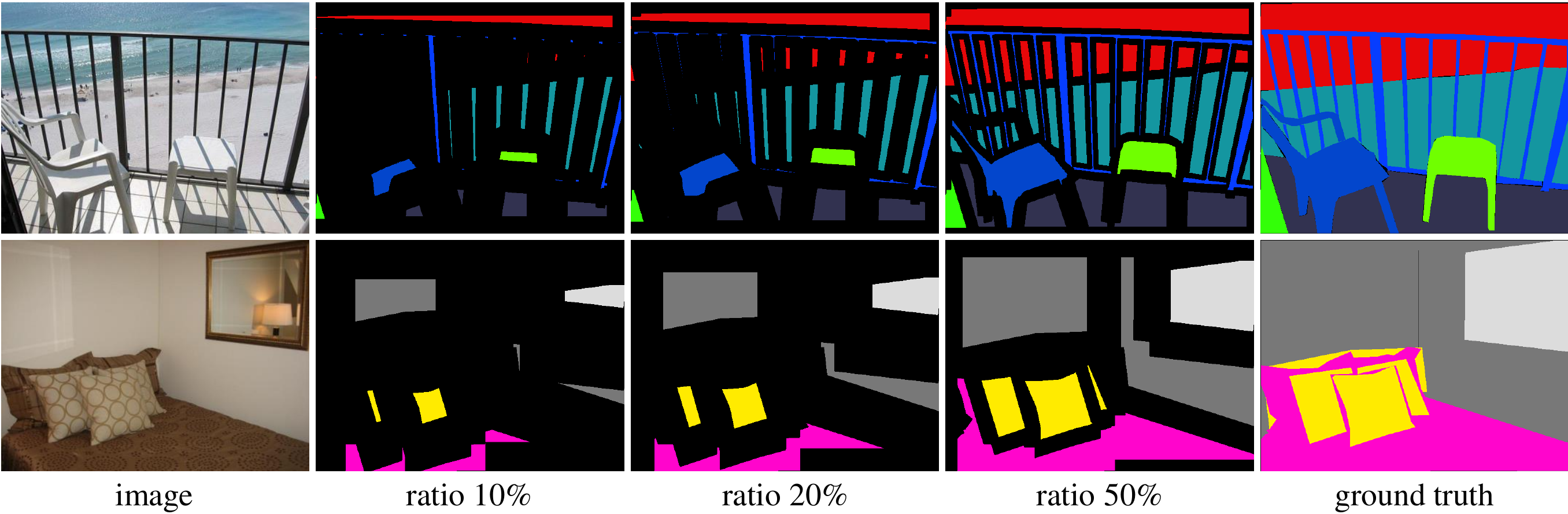}
	\end{center}
	\vspace{-6mm}
	\caption{The block-wise annotations for ADE20k dataset. 
% 	The percentage value denotes the ratio of rest annotations after label removal. The ratio 100\% is the full label of the image.
	}
	\vspace{-4mm}
	\label{Fig:anno_ade}
\end{figure*}

\begin{figure*}[t]
	\begin{center}
		\includegraphics[width=1.\linewidth]{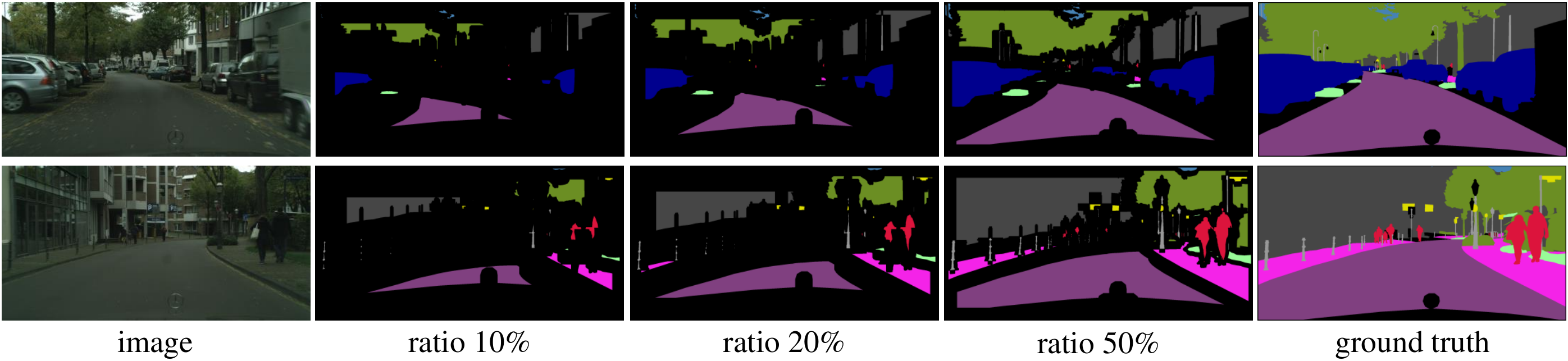}
	\end{center}
	\vspace{-6mm}
	\caption{The block-wise annotations for Cityscapes dataset. 
% 	The percentage value denotes the ratio of rest annotations after label removal. The ratio 100\% is the full label of the image.
	}
	\vspace{-4mm}
	\label{Fig:anno_citys}
\end{figure*}

\begin{figure*}[t]
	\begin{center}
		\includegraphics[width=1.\linewidth]{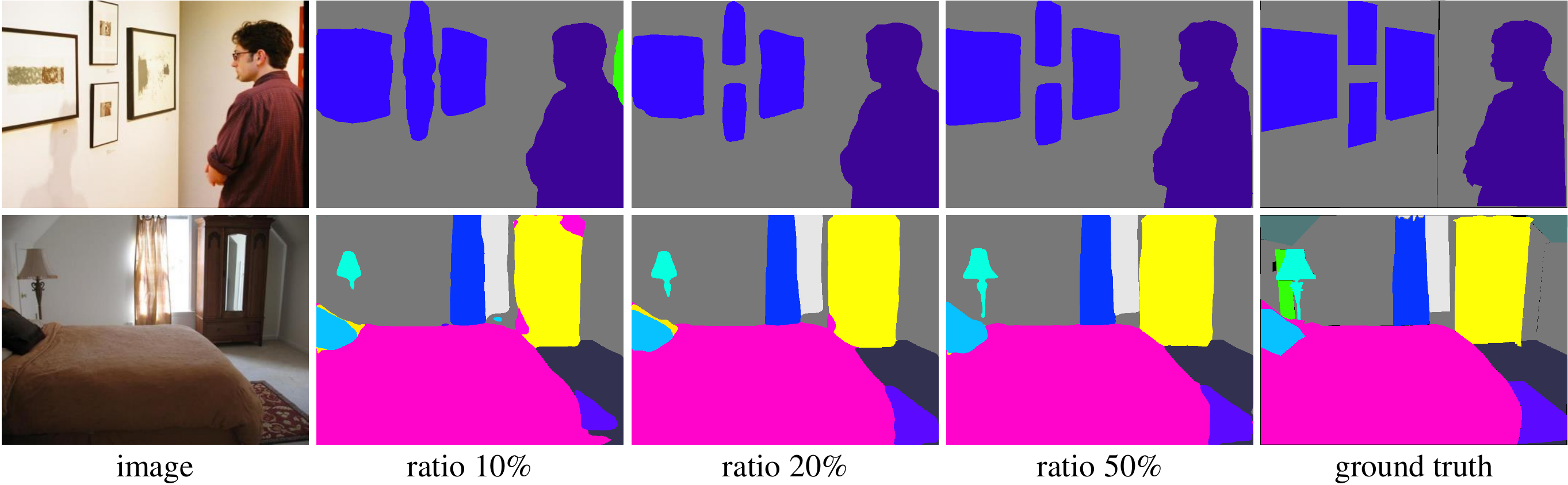}
	\end{center}
	\vspace{-6mm}
	\caption{Qualitative results for the proposed TEL on ADE20k dataset. 
% 	The ratios indicate the segmentation results trained with the corresponding block-wise annotations.
	}
	\vspace{-4mm}
	\label{Fig:res_ade}
\end{figure*}

\begin{figure*}[t]
	\begin{center}
		\includegraphics[width=1.\linewidth]{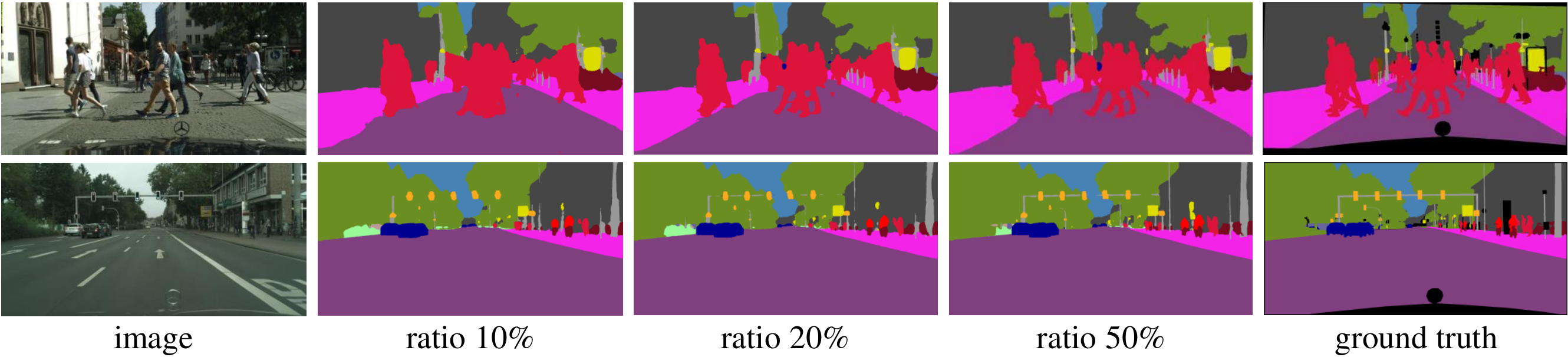}
	\end{center}
	\vspace{-6mm}
	\caption{Qualitative results for the proposed TEL on Cityscapes dataset. 
% 	The ratios indicate the segmentation results trained with the corresponding block-wise annotations.
	}
% 	\vspace{-2mm}
	\label{Fig:res_citys}
\end{figure*}

% \noindent \textbf{Annotations.}
We synthesize the block-wise annotations for ADE20k \cite{zhou2017scene} and Cityscapes \cite{cordts2016cityscapes} datasets. Given the full annotations of the image, we discard the annotations from the semantic boundary to the interior region until the ratio of the rest annotations reaches the preset threshold. The block-wise annotations are generated at 3 levels, including 10\%, 20\%, and 50\% of full annotations. The visualizations of block-wise annotation for the two datasets are respectively illustrated in Fig.~\ref{Fig:anno_ade} and Fig.~\ref{Fig:anno_citys}.
Moreover, qualitative results of our method on the ADE20k and the Cityscapes datasets are illustrated in Fig.~\ref{Fig:res_ade} and Fig.~\ref{Fig:res_citys}, respectively. 
Here, HRNet is selected as the segmentation model.
The proposed approach can be used to train the segmentation model with the annotations of different sparsity.

\end{document}